\documentclass{article}
\usepackage[textwidth=14cm,textheight=24cm]{geometry}
\usepackage{fullname}
\usepackage{dsfont}
\usepackage{MnSymbol}
\usepackage{extarrows}
\usepackage{multirow}
\usepackage{hhline}
\usepackage{url}
\usepackage{latexsym}
\usepackage{graphicx}
\usepackage{tikz-dependency}
\usepackage{ragged2e}
\usepackage{xparse}
\usepackage{float}
\usepackage{graphicx}
\usepackage{wrapfig}
\usepackage{lscape}
\usepackage{rotating}
\usepackage{epstopdf}
\usepackage{color}
\usetikzlibrary{arrows}
\usetikzlibrary{shapes}

\makeatletter
\newsavebox\myboxA
\newsavebox\myboxB
\newlength\mylenA

\newcommand*\xoverline[2][0.75]{%
    \sbox{\myboxA}{$\m@th#2$}%
    \setbox\myboxB\null
    \ht\myboxB=\ht\myboxA%
    \dp\myboxB=\dp\myboxA%
    \wd\myboxB=#1\wd\myboxA
    \sbox\myboxB{$\m@th\overline{\copy\myboxB}$}
    \setlength\mylenA{\the\wd\myboxA}
    \addtolength\mylenA{-\the\wd\myboxB}%
    \ifdim\wd\myboxB<\wd\myboxA%
       \rlap{\hskip 0.5\mylenA\usebox\myboxB}{\usebox\myboxA}%
    \else
        \hskip -0.5\mylenA\rlap{\usebox\myboxA}{\hskip 0.5\mylenA\usebox\myboxB}%
    \fi}
\makeatother

\newcommand{\aptvar}{\mathbf{A}}
\DeclareDocumentCommand\apt{ m g }{%
    {\|#1\IfNoValueF {#2} {;#2}\|}}
\DeclareDocumentCommand\aptanyt{ m g }{%
    {\|#1\IfNoValueF {#2} {;#2}\|^{\textsc{atc}}}}
\DeclareDocumentCommand\aptanys{ m g }{%
    {\|#1\IfNoValueF {#2} {;#2}\|^{\textsc{asc}}}}
\DeclareDocumentCommand\aptany{ m g }{%
    {\|#1\IfNoValueF {#2} {;#2}\|^{\textsc{tc}}}}
\newcommand{\APT}{%
	\textsc{Apt}}
\DeclareDocumentCommand\compapt{ m m g }{%
    {\|#1 ; #2\IfNoValueF {#3} {,#3}\|}}
\DeclareDocumentCommand\comptree{ m g }{%
    {\|#1\IfNoValueF {#2} {;#2}\|}}
\newcommand{\cooccurstree}[2]{%
	\#({#1},{#2})}
\newcommand{\cooccurs}[1]{%
	\#{#1}}
\newcommand{\ctA}[1]{%
	\mbox{\small\sc #1\/}}
\newcommand{\ctB}[2]{%
	\ctA{#1}\!\cdot\!\ctA{#2}}
\newcommand{\ctC}[3]{%
	\ctA{#1}\!\cdot\!\ctA{#2}\!\cdot\!\ctA{#3}}
\newcommand{\ctD}[4]{%
	\ctA{#1}\!\cdot\!\ctA{#2}\!\cdot\!\ctA{#3}\!\cdot\!\ctA{#4}}
\newcommand{\ctE}[5]{%
	\ctA{#1}\!\cdot\!\ctA{#2}\!\cdot\!\ctA{#3}\!\cdot\!\ctA{#4}\!\cdot\!\ctA{#5}}
\newcommand{\dimension}[2]{%
	{#1}\left[#2\right]}
\newcommand{\feat}[2]{%
	\left\langle\,{#1},{#2}\,\right\rangle}
\newcommand{\features}[1]{%
	\mbox{\sc feats}}
\newcommand{\inv}[1]{%
	{\textoverline{#1}}}
\newcommand{\inverse}[1]{%
	{#1}^{-1}}
\newcommand{\lexeme}[2]{%
	\mbox{\em #1\/}\hspace*{-0.8pt}/\hspace*{-0.8pt}\mbox{\scriptsize #2}}
\newcommand{\lexemeA}[2]{%
	\mbox{\em #1\/}}
\newcommand{\merge}{%
	\bigsqcup}
\newcommand{\MIN}{%
	\textsc{int}}

\DeclareDocumentCommand\offapt{ m m g }{%
    {\|#1\IfNoValueF {#3} {;#3}\|{^{#2}}}}

\newcommand{\pmi}[3]{%
	\mbox{\textsc{pmi}}\left({#3},{#1};{#2}\right)}

\newcommand{\pmia}[4]{%
	\mbox{\textsc{pmi}}_{#4}\left({#3},{#1};{#2}\right)}
\newcommand{\sppmi}[3]{%
	\mbox{\textsc{sppmi}}\left({#3},{#1};{#2}\right)}
\newcommand{\PW}[2]{%
	\phi({#1},{#2})}
\newcommand{\reduced}[1]{%
	\downarrow\!({#1})}
\newcommand{\similar}[2]{%
	\mbox{\textsc{sim}}({#1},{#2})}
\newcommand{\SUMMING}{%
	\textsc{uni}}

\newcommand{\tco}[3]{%
	\left\langle {#1},\, {#3},\,{#2}\right\rangle} 
\makeatletter
\newcommand*{\textoverline}[1]{$\overline{\hbox{#1}}\m@th$}
\makeatother
\newcommand{\toggle}[1]{%
	{#1}^{-1}}

\newcommand{\vectorise}[1]{%
	\overrightarrow{#1}}

\newcommand{\weight}[3]{\textsc{w}({#1},\,\feat{#2}{#3})}

\newcommand	   {\pcond}[2]	{p({#1}\,|\,{#2})}

\newcommand     {\set}[1]       {\left\{\,{#1}\,\right\}}

\newcommand     {\setof}[2]     {\left\{\,#1\,\left|\,\,
                                        \mbox{#2}\right.\,\right\}}

\begin{document}

\title{Aligning Packed Dependency Trees: a theory of composition for distributional semantics}
\author{David Weir \and Julie Weeds \and Jeremy Reffin \and Thomas Kober\\[6pt]
Department of Informatics\\ 
University of Sussex\\ 
Falmer, Brighton BN1 9QH\\[5pt]
\{d.j.weir,j.e.weeds,j.p.reffin,t.kober\}@sussex.ac.uk}
\date{}

\maketitle

\begin{abstract}
We present a new framework for compositional distributional semantics in which the distributional contexts of lexemes are expressed in terms of anchored packed dependency trees. We show that these structures have the potential to capture the full sentential contexts of a lexeme and provide a uniform basis for the composition of distributional knowledge in a way that captures both mutual disambiguation and generalization.

\end{abstract}

\section{Introduction}
\label{sec:intro}

This paper addresses a central unresolved issue in distributional semantics: how to model semantic composition. Although there has recently been considerable interest in this problem, it remains unclear what distributional composition actually means.  Our view is that distributional composition is a matter of contextualizing the lexemes being composed. This goes well beyond traditional word sense disambiguation, where each lexeme is assigned one of a fixed number of senses. Our proposal is that composition involves deriving a fine-grained characterization of the distributional meaning of each lexeme in the phrase, where the meaning that is associated with each lexeme is bespoke to that particular context. 

Distributional composition is, therefore, a matter of integrating the meaning of each of the lexemes in the phrase. To achieve this we need  a structure within which all of the lexemes' semantics can be overlaid. Once this is done, the lexemes can collectively agree on the semantics of the phrase, and in so doing, determine the semantics that they have in the context of that phrase. Our process of composition thus creates a single structure that encodes contextualized representations of every lexeme in the phrase.

The (uncontextualized) distributional knowledge of a lexeme is typically formed by aggregating distributional features across \emph{all} uses of the lexeme found within the corpus, where distributional features arise from co-occurrences found in the corpus. The distributional features of a lexeme are associated with weights that encode the strength of that feature. Contextualization involves inferring adjustments to these weights to reflect the  context in which the lexeme is being used. The weights of distributional features that don't fit the context are reduced, while the weight of those features that are compatible with the context can be boosted.

As an example, consider how we contextualize the distributional features of the word \emph{wooden} in the context of the phrase \emph{wooden floor}. The uncontextualized representation of \emph{wooden}  presumably includes distributional features associated with  different uses, for example \emph{The director fired the wooden actor} and \emph{I sat on the wooden chair}. So, while we may have observed in a corpus that it is plausible for the adjective \emph{wooden} to modify \emph{floor}, \emph{table}, \emph{toy}, \emph{actor} and \emph{voice},  in the specific context of the phrase \emph{wooden floor}, we need to find a way to down-weight the  distributional features of being something that can modify \emph{actor} and \emph{voice}, while up-weighting the distributional features of being something that can modify \emph{table} and \emph{toy}. 

In the example above we considered so-called first-order distributional features; these involve a single dependency relation, e.g. an adjective modifying a noun. Similar inferences can also be made with respect to distributional features that involve higher-order grammatical dependencies\protect{\footnote{Given some dependency tree, a $k$-th order dependency holds between two lexemes (nodes) in the tree when the path between the two lexemes has length $k$.}}. For example, suppose that we have observed that a noun that \emph{wooden} modifies (e.g. \emph{actor}) can be the direct object of the verb \emph{fired}, as in \emph{The director fired the wooden actor}. We want this distributional feature of \emph{wooden} to be down-weighted in the distributional representation of \emph{wooden} in the context of \emph{wooden table}, since things made of wood do not typically lose their job. 

In addition to specialising the distributional representation of \emph{wood} to reflect the context \emph{wooden floor}, the distributional representation of \emph{floor} should also be refined, down-weighting distributional features arising in contexts such as \emph{Prices fell through the floor}, while up-weighting distributional features arising in contexts such as \emph{I polished the concrete floor}.

In our example, some of the distributional features of \emph{wooden}, in particular, those to do with the noun that this sense of \emph{wooden} could modify, are \textbf{internal} to the phrase \emph{wooden floor} in the sense that they are alternatives to one of the words in the phrase. Although it is specifically a \emph{floor} that is \emph{wooden}, our proposal is that the contextualized representation of \emph{wooden} should  recognise that it is plausible that nouns such as \emph{chair} and \emph{toy} could be modified by the particular sense of \emph{wooden} that is being used. The remaining distributional features are \textbf{external} to the phrase. For example, the verb \emph{mop} could be an external feature, since things that can be modified by \emph{wooden} can be the direct object of \emph{mop}.  The external features of \emph{wooden} and \emph{floor} with respect to the phrase \emph{wooden floor} provide something akin to the traditional interpretation of the distributional semantics of the phrase, i.e. a representation of those (external) contexts in which this phrase can occur.

While internal features are, in a sense, inconsistent with the specific semantics of the phrase, they provide a way to embellish the characterization of the distributional meaning of the lexemes in the phrase. Recall that our goal is to infer a rich and fine-grained representation of the contextualized distributional meaning of each of the lexemes in the phrase.

Having introduced the proposal that distributional composition should be viewed as a matter of contextualization, the question arises as to how to realise this conception. Since each lexeme in the phrase needs to be able to contribute to the contextualization of the other lexemes in the phrase, we need to be able to align what we know about each of the lexeme's distributional features so that this can be achieved. The problem is that the uncontextualized distributional knowledge associated with the different lexemes in the phrase take a different perspective on the feature space.  To overcome this we need to: (a)~provide a way of structuring the distributional feature space, which we do by typing distributional features with dependency paths; and (b)~find a way to systematically modify the perspective that each lexeme has on this structured feature space in such a way that they are all aligned with one another.

Following \namecite{Baroni_2010b}, we use typed dependency relations as the bases for our distributional features, and following \namecite{Pado_2007}, we include higher-order dependency relations in this space. However, in contrast to previous proposals, the higher order dependency relations provides structure to the space which is crucial to our definition of composition. Each co-occurrence associated with a lexeme such as \emph{wooden} is typed by the path in the dependency tree that connects the lexeme \emph{wooden} with the co-occurring lexeme, e.g. \emph{fired}. This allows us to encode a lexeme's distributional knowledge with a hierarchical structure that we call an Anchored Packed Dependency Tree~({\APT}). As we show, this data structure provides a way for us to align the distributional knowledge of the lexemes that are being composed in such a way that the inferences needed to achieve contextualization can be implemented.

\section{The Distributional Lexicon}
\label{sec:lexicon}

In this section, we begin the formalisation of our proposal by describing the distributional lexicon: a collection of entries that characterize the distributional semantics of lexemes. Table~\ref{fig:notation} provides a summary of the notation that we are using. 

Let $V$ be a finite alphabet of lexemes\footnote{There is no reason why lexemes could not include multi-word phrases tagged with an appropriate part of speech.}, where each lexeme is assumed to incorporate a part-of-speech tag; let $R$ be a finite alphabet of grammatical dependency relations; and let $T_{V,R}$ be the set of dependency trees where every node is labeled with a member of $V$, and every directed edge is labeled with an element of $R$. Figure~\ref{fig:deptrees} shows eight examples of dependency trees.

\begin{table}[2pt]
\centering
\begin{displaymath}
	\begin{array}{|c@{\quad}l@{\quad}|}\hline
	\mbox{\bf Notation} & \mbox{\bf Description}\\\hline
	\hspace*{3cm}&\\[-4pt]
	V  & 
		\mbox{a finite set of lexemes}\\[2pt]
	w & 
		\mbox{a lexeme}\\[2pt]
	R & 
		\mbox{a finite set of dependency tree edge labels}\\[2pt]
	r & 
		\mbox{an element of $R$}\\[2pt]
	\xoverline{R} & 
		\mbox{a finite set of inverse dependency tree edge labels}\\[2pt]
	\xoverline{r} & 
		\mbox{an element of \xoverline{R}}\\[2pt]
	x &
		\mbox{an element of $R\cup\xoverline{R}$ }\\[2pt]
	T_{V,R} & 
		\mbox{the dependency trees over lexemes $V$ and dependencies $R$}\\[2pt]
	t & 
		\mbox{a dependency tree}\\[2pt]
	\tau & 
		\mbox{a co-occurrence type (path)}\\[2pt]
	\inverse{\tau} & 
		\mbox{the inverse (reverse) of path $\tau$}\\[2pt]
	\tco{w}{w'}{\tau} &
		\mbox{the co-occurrence of $w$ with $w'$ with co-occurrence type $\tau$}\\[2pt]
	C & 
		\mbox{a corpus of (observed) dependency trees}\\[2pt]
	\reduced{\tau} &
		\mbox{the co-occurrence type produced by reducing $\tau$}\\[2pt]
	\cooccurstree{\tco{w}{w'}{\tau}}{t} &
		\mbox{number of occurrences of $\tco{w}{w'}{\tau}$ in $t$}\\[2pt]
	\cooccurs{\tco{w}{w'}{\tau}} &
		\mbox{number of occurrences of $\tco{w}{w'}{\tau}$ in the corpus}\\[2pt]
	\apt{w} & 
		\mbox{the (uncontextualized) {\APT} for $w$}\\[2pt]
	\aptvar & 
		\mbox{an {\APT}}\\[2pt]
	\apt{w}(\tau,w') & 
		\mbox{the weight for $w'$ in $\apt{w}$ at node for co-occurrence type $\tau$}\\[2pt]
	\apt{w}(\tau) & 
		\mbox{the node (weighted lexeme multiset) in $\apt{w}$ for co-occurrence type $\tau$}\\[2pt]
	\features{C} &
		\mbox{the set of all distributional features arising in $C$}\\[2pt]
	\feat{\tau}{w} &
		\mbox{a distributional feature in vector space}\\[2pt]
	\weight{w}{\tau}{w'} & 
		\mbox{the weight of the distributional feature $\feat{\tau}{w'}$ of lexeme $w$}\\[2pt]
	\vectorise{\apt{w}} &
		\mbox{the vector representation of the {\APT} $\apt{w}$}\\[2pt] 
	\similar{\apt{w_1}}{\apt{w_2}} &
		\mbox{the distributional similarity of $\apt{w_1}$ and $\apt{w_2}$}\\[2pt]
	\offapt{w}{\delta} & 
		\mbox{the {\APT} $\apt{w}$ that has been offset by $\delta$}\\[2pt]
	\mbox{$\comptree{t}$}  &
		\mbox{the composed {\APT} for the tree $t$}\\[2pt]
	\compapt{w}{t} &
		\mbox{the {\APT} for $w$ when contextualized by $t$}\\[2pt]
	\merge\set{\aptvar_1,\ldots,\aptvar_n} &
		\mbox{the result of merging aligned {\APT}s in $\set{\aptvar_1,\ldots,\aptvar_n}$}\\[2pt]		\hline
	\end{array}
\end{displaymath}
\caption{Summary of notation}
\label{fig:notation}
\end{table}

\subsection{Typed Co-occurrences}
\label{sec:typedcooccurrences}

When two lexemes $w$ and $w'$  co-occur in a dependency tree%
\footnote{In order to avoid over-complicating our presentation, when possible, we do not  distinguish between a node in a dependency tree and the lexeme that appears at that node.}
in $t\in T_{V,R}$, we represent this co-occurrence as a triple $\tco{w}{w'}{\tau}$ where $\tau$ is a string that encodes the \textbf{co-occurrence type} of this co-occurrence, capturing the syntactic relationship that holds between these occurrences of the two lexemes. In particular, $\tau$ encodes the sequence of dependencies that lie along the path in $t$ between the occurrences of $w$ and $w'$ in $t$. In general, a path from $w$ to $w'$ in $t$  initially travels up towards the root of $t$ (against the directionality of the dependency edges) until an ancestor of $w'$ is reached. It  then travels down the tree to $w'$ (following the directionality of the dependencies).  The string $\tau$ must, therefore, not only encode the sequence of dependency relations appearing along the path, but also whether each edge is traversed in a forward or backward direction. In particular, given the path $\langle v_0,\ldots,v_k\rangle$ in $t$, where $k>0$,   $w$ labels $v_0$  and $w'$ labels $v_k$, the string $\tau= x_1\ldots x_k$  encodes the co-occurrence type associated with this path as follows:
\begin{itemize}
	\item if the edge connecting $v_{i-1}$ and $v_{i}$ runs from $v_{i-1}$ to $v_{i}$ and is labeled by $r$ then $x_i=r$; and 
	\item if the edge connecting $v_{i-1}$ and $v_{i}$ runs from $v_{i}$ to $v_{i-1}$ and is labeled by $r$ then $x_i=\xoverline{r}$. 
\end{itemize}
Hence, co-occurrence types are strings in $\xoverline R^*R^*$, where $\xoverline R = \setof{\xoverline r}{$r\in R$}$.  

It is useful to be able to refer to the \textbf{order} of a co-occurrence type, where this simply refers to the length of the dependency path.  It is also convenient to be able to refer to the inverse of a co-occurrence type. This can be thought of as the same path, but traversed in the reverse direction.  To be precise, given the co-occurrence type $\tau=x_1\cdot\ldots\cdot x_n$ where each $x_i\in R\cup \xoverline{R}$ for $1\le i\le n$, the inverse of $\tau$, denoted $\inverse{\tau}$, is the path $\toggle{x_n}\cdot\ldots\cdot\toggle{x_1}$ where $\toggle{r}=\xoverline r$ and $\toggle{\xoverline r}=r$ for $r\in R$. For example, the inverse of 
$\ctC{\inv{amod}}{dobj}{nsubj}$ is $\ctC{\inv{nsubj}}{\inv{dobj}}{amod}$. 

\begin{figure}
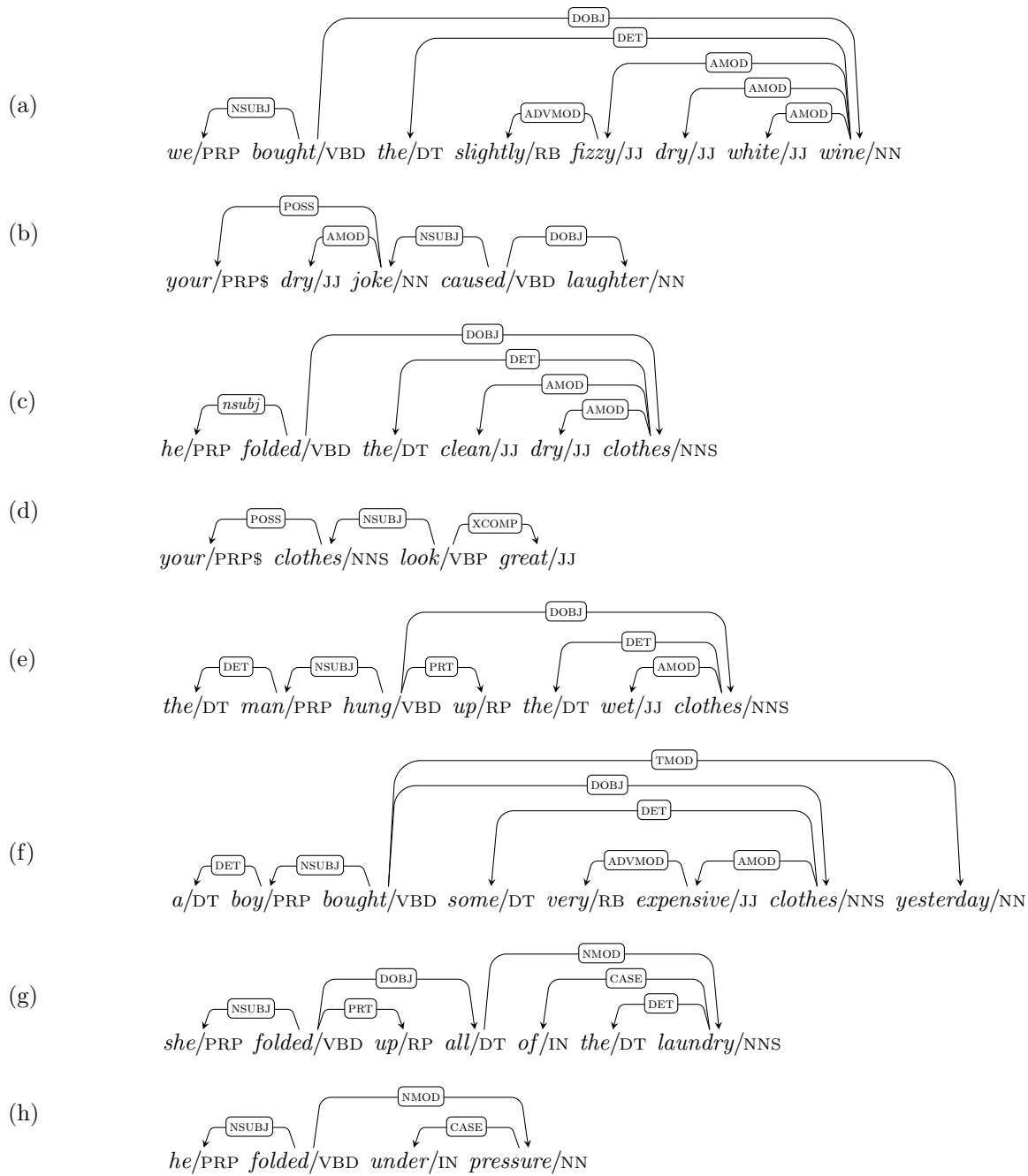

\begin{dependency}
\begin{deptext}[column sep=.0cm, row sep=.1ex]
\lexeme{we}{PRP} \& \lexeme{bought}{VBD}  \& \lexeme{the}{DT} \& \lexeme{slightly}{RB} \& \lexeme{fizzy}{JJ} \& \lexeme{dry}{JJ} \& \lexeme{white}{JJ} \& \lexeme{wine}{NN} \\
   \end{deptext}
   \depedge{2}{1}{\ctA{nsubj}}
   \depedge[edge unit distance=2ex]{2}{8}{\ctA{dobj}}
   \depedge[edge unit distance=2ex]{8}{3}{\ctA{det}}
   \depedge[edge unit distance=2.5ex]{5}{4}{\ctA{advmod}}
   \depedge[edge unit distance=2.5ex]{8}{5}{\ctA{amod}}
   \depedge[edge unit distance=2.5ex]{8}{6}{\ctA{amod}}
   \depedge[edge unit distance=2.5ex]{8}{7}{\ctA{amod}}
   \node (silly1) [above left of = \wordref{1}{1}, xshift = -2cm] {(a)};  
\end{dependency}\\[8pt]
\begin{dependency}
   \begin{deptext}[column sep=.0cm, row sep=.1ex]
      \lexeme{your}{PRP\$}  \& \lexeme{dry}{JJ} \& \lexeme{joke}{NN} \& \lexeme{caused}{VBD}  \& \lexeme{laughter}{NN}\\
   \end{deptext}
   \depedge{3}{1}{\ctA{poss}}
   \depedge{3}{2}{\ctA{amod}}
   \depedge{4}{3}{\ctA{nsubj}}
   \depedge{4}{5}{\ctA{dobj}}
   \node (silly1) [above left of = \wordref{1}{1}, xshift = -2.2cm] {(b)};
\end{dependency}\\[8pt]
\begin{dependency}
   \begin{deptext}[column sep=.0cm, row sep=.1ex]
      \lexeme{he}{PRP} \& \lexeme{folded}{VBD} \& \lexeme{the}{DT} \& \lexeme{clean}{JJ} \& \lexeme{dry}{JJ} \& \lexeme{clothes}{NNS} \\
   \end{deptext}
   \depedge{2}{1}{\emph{nsubj}}
   \depedge[edge unit distance=2.5ex]{2}{6}{\ctA{dobj}}
   \depedge[edge unit distance=2.5ex]{6}{3}{\ctA{det}}
   \depedge[edge unit distance=2.5ex]{6}{4}{\ctA{amod}}
   \depedge[edge unit distance=2.5ex]{6}{5}{\ctA{amod}}
   \node (silly1) [above left of = \wordref{1}{1}, xshift = -1.9cm] {(c)};
\end{dependency}\\[8pt]
\begin{dependency}
   \begin{deptext}[column sep=.0cm, row sep=.1ex]
      \lexeme{your}{PRP\$} \& \lexeme{clothes}{NNS} \& \lexeme{look}{VBP} \& \lexeme{great}{JJ} \\
   \end{deptext}
   \depedge[edge unit distance=2.5ex]{2}{1}{\ctA{poss}}
   \depedge[edge unit distance=2.5ex]{3}{2}{\ctA{nsubj}}
   \depedge[edge unit distance=2ex]{3}{4}{\ctA{xcomp}}
   \node (silly1) [above left of = \wordref{1}{1}, xshift = -2.1cm] {(d)};
\end{dependency}\\[8pt]
\begin{dependency}
   \begin{deptext}[column sep=.0cm, row sep=.1ex]
    \lexeme{the}{DT} \& \lexeme{man}{PRP} \& \lexeme{hung}{VBD} \& \lexeme{up}{RP} \& \lexeme{the}{DT} \& \lexeme{wet}{JJ} \& \lexeme{clothes}{NNS} \\
   \end{deptext}
   \depedge[edge unit distance=2.5ex]{2}{1}{\ctA{det}}
   \depedge[edge unit distance=2.5ex]{3}{2}{\ctA{nsubj}}
   \depedge[edge unit distance=2.5ex]{3}{4}{\ctA{prt}}
   \depedge[edge unit distance=2ex]{3}{7}{\ctA{dobj}}
   \depedge[edge unit distance=2.5ex]{7}{5}{\ctA{det}}
   \depedge[edge unit distance=2.5ex]{7}{6}{\ctA{amod}}
   \node (silly1) [above left of = \wordref{1}{1}, xshift = -1.9cm] {(e)};
\end{dependency}\\[8pt]
\begin{dependency}
   \begin{deptext}[column sep=.0cm, row sep=.1ex]
      \lexeme{a}{DT} \& \lexeme{boy}{PRP} \& \lexeme{bought}{VBD}  \& \lexeme{some}{DT} \& \lexeme{very}{RB} \& \lexeme{expensive}{JJ} \& \lexeme{clothes}{NNS} \& \lexeme{yesterday}{NN}\\
   \end{deptext}
   \depedge[edge unit distance=2ex]{2}{1}{\ctA{det}}
   \depedge[edge unit distance=2ex]{3}{2}{\ctA{nsubj}}
   \depedge[edge unit distance=2.5ex]{7}{4}{\ctA{det}}
   \depedge[edge unit distance=2.5ex]{6}{5}{\ctA{advmod}}
   \depedge[edge unit distance=2.5ex]{7}{6}{\ctA{amod}}
   \depedge[edge unit distance=2.5ex]{3}{7}{\ctA{dobj}}
   \depedge[edge unit distance=2.5ex]{3}{8}{\ctA{tmod}}
   \node (silly1) [above left of = \wordref{1}{1}, xshift = -1.9cm] {(f)};
\end{dependency}\\[8pt]
\begin{dependency}
   \begin{deptext}[column sep=.0cm, row sep=.1ex]
      \lexeme{she}{PRP} \& \lexeme{folded}{VBD} \& \lexeme{up}{RP} \& \lexeme{all}{DT} \& \lexeme{of}{IN} \& \lexeme{the}{DT}  \& \lexeme{laundry}{NNS} \\
   \end{deptext}
   \depedge[edge unit distance=2ex]{2}{1}{\ctA{nsubj}}
   \depedge[edge unit distance=2ex]{2}{3}{\ctA{prt}}
   \depedge[edge unit distance=2.5ex]{2}{4}{\ctA{dobj}}
   \depedge[edge unit distance=2.5ex]{7}{5}{\ctA{case}}
   \depedge[edge unit distance=2.5ex]{7}{6}{\ctA{det}}
   \depedge[edge unit distance=2.5ex]{4}{7}{\ctA{nmod}}
   \node (silly1) [above left of = \wordref{1}{1}, xshift = -2cm] {(g)};
\end{dependency}\\[8pt]
\begin{dependency}
   \begin{deptext}[column sep=.0cm, row sep=.1ex]
      \lexeme{he}{PRP} \& \lexeme{folded}{VBD} \& \lexeme{under}{IN}  \& \lexeme{pressure}{NN}\\
   \end{deptext}
   \depedge[edge unit distance=2ex]{2}{1}{\ctA{nsubj}}
   \depedge[edge unit distance=2ex]{4}{3}{\ctA{case}}
   \depedge[edge unit distance=2.5ex]{2}{4}{\ctA{nmod}}
   \node (silly1) [above left of = \wordref{1}{1}, xshift = -2cm] {(h)};
\end{dependency}
\caption{A small corpus of dependency trees.}
\label{fig:deptrees}
\end{figure}

The following typed co-occurrences for the lexeme \lexeme{white}{JJ} arise in the tree shown in Figure~\ref{fig:deptrees}(a). 
\begin{displaymath}
\begin{array}{l@{\qquad}l}

\tco{\lexeme{white}{JJ}}{\lexeme{we}{PRP}}{\ctC{\inv{amod}}{\inv{dobj}}{nsubj}} &
\tco{\lexeme{white}{JJ}}{\lexeme{fizzy}{JJ}}{\ctB{\inv{amod}}{amod}}\\[5pt]
\tco{\lexeme{white}{JJ}}{\lexeme{bought}{VBD}}{\ctB{\inv{amod}}{\inv{dobj}}} &
\tco{\lexeme{white}{JJ}}{\lexeme{dry}{JJ}}{\ctB{\inv{amod}}{amod}} \\[5pt]
\tco{\lexeme{white}{JJ}}{\lexeme{the}{DT}}{\ctB{\inv{amod}}{det}} &
\tco{\lexeme{white}{JJ}}{\lexeme{white}{JJ}}{\epsilon} \\[5pt]
\tco{\lexeme{white}{JJ}}{\lexeme{slightly}{RB}}{\ctC{\inv{amod}}{amod}{advmod}} &
\tco{\lexeme{white}{JJ}}{\lexeme{wine}{NN}}{\ctA{\inv{amod}}}\end{array}
\end{displaymath}

Notice that we have included the co-occurrence $\tco{\lexeme{white}{JJ}}{\lexeme{white}{JJ}}{\epsilon}$. This gives a uniformity to our typing system that simplifies the formulation of distributional composition in Section~\ref{sec:composition}, and leads to the need for a refinement to our co-occurrence type encodings. Since we permit paths that traverse both forwards and backwards along the same dependency, e.g. in the co-occurrence $\tco{\lexeme{white}{JJ}}{\lexeme{dry}{JJ}}{\ctB{\inv{amod}}{amod}}$, it is logical to consider $\tco{\lexeme{white}{JJ}}{\lexeme{dry}{JJ}}{\ctD{\inv{amod}}{\inv{dobj}}{dobj}{amod}}$ a valid co-occurrence. However, in line with our decision to include $\tco{\lexeme{white}{JJ}}{\lexeme{white}{JJ}}{\epsilon}$ rather than $\tco{\lexeme{white}{JJ}}{\lexeme{white}{JJ}}{\ctB{\inv{amod}}{amod}}$, all co-occurrence types are canonicalized through a dependency cancellation process in which adjacent, complementary dependencies are cancelled out. In particular, all occurrences within the string of either $r\xoverline{r}$ or $\xoverline{r}r$ for  $r\in R$ are replaced with $\epsilon$, and this process is repeated until no further reductions are possible.

The reduced co-occurrence type produced from $\tau$ is denoted $\reduced{\tau}$, and defined as follows:
\begin{equation}
\reduced{\tau}=\left\{\begin{array}{l@{\quad}l}
	\reduced{\tau_1\tau_2} &\mbox{if $\tau=\tau_1\, r\,\xoverline{r}\,\tau_2$ or $\tau=\tau_1\,\xoverline{r}\,r\,\tau_2$ for some $r\in R$}\\[2pt]
	\tau & \mbox{otherwise}
\end{array}\right.
\label{eq:reduced}
\end{equation}
For the remainder of the paper, we  only consider reduced co-occurrence types when associating a type with a co-occurrence. 
 
Given a tree $t\in T_{V,R}$, lexemes $w$ and $w'$ and reduced co-occurrence type $\tau$, the number of times that the co-occurrence $\tco{w}{w'}{\tau}$ occurs in $t$ is denoted $\cooccurstree{\tco{w}{w'}{\tau}}{t}$, and, given some corpus $C$ of dependency trees, the sum of all $\cooccurstree{\tco{w}{w'}{\tau}}{t}$ across all $t\in C$ is denoted $\cooccurs{\tco{w}{w'}{\tau}}$. Note that in order to simplify our notation, the dependence on the corpus $C$ is not expressed in our notation.

It is common to use alternatives to raw counts in order to capture the strength of each distributional feature. A variety of alternatives are considered during the experimental work presented in Section~\ref{sec:empirical}. Among the options we have considered are probabilities and various versions of positive pointwise mutual information. While, in practice, the precise method for weighting features is of practical importance, it is not an intrinsic part of the theory that this paper is introducing. In the exposition below we denote the weight of the distributional feature $\feat{\tau}{w'}$ of the lexeme $w$ with the expression $\weight{w}{\tau}{w'}$. 
\subsection{Anchored Packed Trees}
\label{sec:APTs}

Given a dependency tree corpus  $C\subset T_{V,R}$ and a lexeme $w\in V$, we are interested in capturing the aggregation of all  distributional contexts  of $w$ in $C$ within a single structure. We achieve this with what we call an \textbf{Anchored Packed Tree}~(\APT). {\APT}s are central to the proposals in this paper: not only can they  be used to encode the aggregate of all distributional features of a lexeme over a corpus of dependency trees, but they can also be used to express the  distributional features of a lexeme that has been contextualized within some dependency tree (see Section~\ref{sec:composition}). 

The {\APT} for $w$ given $C$, is denoted $\apt{w}$, and referred to as the \textbf{elementary {\APT}} for $w$. Below, we describe a tree-based interpretation of $\apt{w}$, but in the first instance we  define it as a mapping from pairs $(\tau,w')$ where $\tau\in\xoverline{R}^*R^*$ and $w'\in V$, such that $\apt{w}(\tau,w')$ gives the weight of the typed co-occurrence $\tco{w}{w'}{\tau}$ in the corpus $C$. It is nothing more than those components of the weight function that specify the weights of distributional features of $w$. In other words, for each $\tau\in\xoverline{R}^*R^*$ and $w'\in V$:
\begin{equation}
\apt{w}(\tau,w') = 
\weight{w}{\tau}{w'}
\label{eq:aptn}
\end{equation}
The restriction of $\apt{w}$ to co-occurrence types that are at most order $k$ is referred to as a $k$-th order {\APT}. The \textbf{distributional lexicon} derived from a corpus $C$ is a collection of lexical entries where the entry for the lexeme $w$ is the elementary {\APT} $\apt{w}$. 

Formulating {\APT}s as functions simplifies the definitions that appear below. However, since an {\APT} encodes co-occurrences that are aggregated over a set of dependency trees, they can also be interpreted as having a tree structure. In our tree-based interpretation of {\APT}s, nodes are associated with weighted multisets of lexemes. In particular,  $\apt{w}(\tau)$ is thought of as a node that is associated with the weighted lexeme multiset in which the weight of $w'$ in the multiset is $\apt{w}(\tau,w')$. We refer to the node $\apt{w}(\epsilon)$ as the \textbf{anchor} of the {\APT} $\apt{w}$. 

\begin{figure}
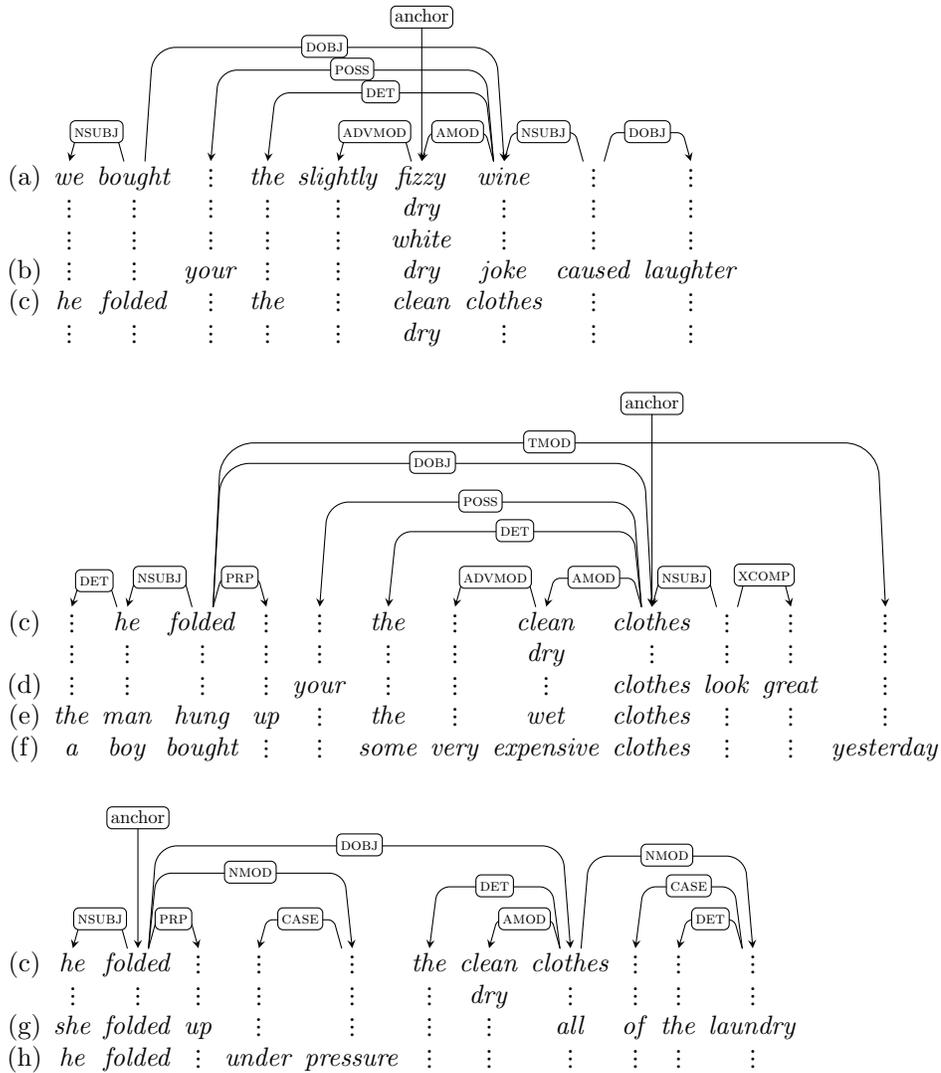

\begin{dependency}
\begin{deptext}[column sep=.0cm, row sep=.0ex]
(a)\&
  \lexemeA{we}{PRP}\&
    \lexemeA{bought}{VBD}\& 
      {$\vdots$}\&
        \lexemeA{the}{DT}\&
          \lexemeA{slightly}{RB}\&
            \lexemeA{fizzy}{JJ}\&
              \lexemeA{wine}{NN}\&
                {$\vdots$}\&
                  {$\vdots$}\\
\&
  {$\vdots$}\&
    {$\vdots$}\&
      {$\vdots$}\&
        {$\vdots$}\&
          {$\vdots$}\&
            \lexemeA{dry}{JJ}\&
              {$\vdots$}\&
                {$\vdots$}\&
                  {$\vdots$}\\
\&
  {$\vdots$}\&
    {$\vdots$}\&
      {$\vdots$}\&
        {$\vdots$}\&
          {$\vdots$}\&
            \lexemeA{white}{JJ}\&
              {$\vdots$}\&
                {$\vdots$}\&
                  {$\vdots$}\\
(b)\&
  {$\vdots$}\&
    {$\vdots$}\&
      \lexemeA{your}{PRP\$}\&
        {$\vdots$}\&
          {$\vdots$}\&
            \lexemeA{dry}{JJ}\&
              \lexemeA{joke}{NN}\&
                \lexemeA{caused}{VBD}\&
                  \lexemeA{laughter}{NN}\\
(c)\&
  \lexemeA{he}{PRP}\&
    \lexemeA{folded}{VBD}\&
      {$\vdots$}\&
        \lexemeA{the}{DT}\&
          {$\vdots$}\&
            \lexemeA{clean}{JJ}\&
              \lexemeA{clothes}{NNS}\&
                {$\vdots$}\&
                  {$\vdots$}\\
\&
  {$\vdots$}\&
    {$\vdots$}\&
      {$\vdots$}\&
        {$\vdots$}\&
          {$\vdots$}\&
           \lexemeA{dry}{JJ}\&
              {$\vdots$}\&
                {$\vdots$}\&
                  {$\vdots$}\\          
  \&
    \&
      \&
        \&
          \&
            \&
                \&
                    \&
                      \&
                              \&
                                \&
                                  \&\\[4pt]
\end{deptext}
\deproot[edge unit distance=3.5ex]{7}{anchor}
\depedge[edge unit distance=2.5ex]{3}{2}{\ctA{nsubj}}
\depedge[edge unit distance=2ex]{3}{8}{\ctA{dobj}}
\depedge[edge unit distance=2ex]{8}{4}{\ctA{poss}}
\depedge[edge unit distance=2ex]{8}{5}{\ctA{det}}
\depedge[edge unit distance=2.5ex]{7}{6}{\ctA{advmod}}
\depedge[edge unit distance=2.5ex]{8}{7}{\ctA{amod}}
\depedge[edge unit distance=2.5ex]{9}{8}{\ctA{nsubj}}
\depedge[edge unit distance=2.5ex]{9}{10}{\ctA{dobj}}
\end{dependency}
\begin{dependency}
\begin{deptext}[column sep=.0cm, row sep=.0ex]
(c)\&
  {$\vdots$}\&
    \lexemeA{he}{PRP}\&
      \lexemeA{folded}{VDB}\&
        {$\vdots$}\&
              {$\vdots$}\&
                \lexemeA{the}{DT}\&
                  {$\vdots$}\&
                    \lexemeA{clean}{JJ}\&
                      \lexemeA{clothes}{NNS}\&
                        {$\vdots$}\&
                            {$\vdots$}\&
                                    {$\vdots$}\&\\
\&
  {$\vdots$}\&
    {$\vdots$}\&
      {$\vdots$}\&
        {$\vdots$}\&
              {$\vdots$}\&
                {$\vdots$}\&
                  {$\vdots$}\&
                    \lexemeA{dry}{JJ}\&
                      {$\vdots$}\&
                        {$\vdots$}\&
                            {$\vdots$}\&
                                    {$\vdots$}\&\\
(d)\&
  {$\vdots$}\&
    {$\vdots$}\&
      {$\vdots$}\&
        {$\vdots$}\&
              \lexemeA{your}{PRP\$}\& 
                {$\vdots$}\&
                  {$\vdots$}\&
                    {$\vdots$}\&
                      \lexemeA{clothes}{NNS}\&
                        \lexemeA{look}{VBP}\&
                            \lexemeA{great}{JJ}\&
                                    {$\vdots$}\\  
(e)\&
  \lexemeA{the}{DT}\& 
    \lexemeA{man}{NN}\&
      \lexemeA{hung}{VBD}\&
        \lexemeA{up}{RP}\&
              {$\vdots$}\&
                \lexemeA{the}{DT}\&
                  {$\vdots$}\&
                    \lexemeA{wet}{JJ}\&
                      \lexemeA{clothes}{NNS}\&
                        {$\vdots$}\&
                            {$\vdots$}\&
                                    {$\vdots$}\\  
(f)\&
  \lexemeA{a}{DT}\&
    \lexemeA{boy}{NN}\&
      \lexemeA{bought}{VBD}\&
        {$\vdots$}\&
              {$\vdots$}\&
                \lexemeA{some}{DT}\&
                  \lexemeA{very}{RB}\&
                    \lexemeA{expensive}{JJ}\&
                      \lexemeA{clothes}{NNS}\&
                        {$\vdots$}\&
                            {$\vdots$}\&
                                    \lexemeA{yesterday}{NN}\\
  \&
    \&
      \&
        \&
          \&
            \&
                \&
                    \&
                      \&
                              \&
                                \&
                                  \&\\[4pt]
\end{deptext}
 \deproot[edge unit distance=4.8ex]{10}{anchor}
 \depedge[edge unit distance=2.2ex]{3}{2}{\ctA{det}}
 \depedge[edge unit distance=2.5ex]{4}{3}{\ctA{nsubj}}
 \depedge[edge unit distance=2.5ex]{4}{5}{\ctA{prp}}
 \depedge[edge unit distance=2.3ex]{10}{6}{\ctA{poss}}
 \depedge[edge unit distance=2.2ex]{10}{7}{\ctA{det}}
 \depedge[edge unit distance=2.5ex]{9}{8}{\ctA{advmod}}
 \depedge[edge unit distance=2.5ex]{10}{9}{\ctA{amod}}
 \depedge[edge unit distance=2.5ex]{11}{10}{\ctA{nsubj}}
 \depedge[edge unit distance=2.1ex]{4}{10}{\ctA{dobj}}
 \depedge[edge unit distance=2.7ex]{11}{12}{\ctA{xcomp}}
 \depedge[edge unit distance=1.6ex]{4}{13}{\ctA{tmod}}
\end{dependency} 
\begin{dependency}
\begin{deptext}[column sep=.0cm, row sep=.0ex]
(c)\&
    \lexemeA{he}{PRP}\&
      \lexemeA{folded}{VDB}\&
        {$\vdots$}\&
          {$\vdots$}\&
            {$\vdots$}\&
                \lexemeA{the}{DT}\&
                    \lexemeA{clean}{JJ}\&
                      \lexemeA{clothes}{NNS}\&
                              {$\vdots$}\&
                                {$\vdots$}\&
                                  {$\vdots$}\\
\&
    {$\vdots$}\&
      {$\vdots$}\&
        {$\vdots$}\&
          {$\vdots$}\&
            {$\vdots$}\&
                {$\vdots$}\&
                    \lexemeA{dry}{JJ}\&
                      {$\vdots$}\&
                              {$\vdots$}\&
                                {$\vdots$}\&
                                  {$\vdots$}\\
(g)\&
    \lexemeA{she}{PRP}\&
      \lexemeA{folded}{VBD}\&
        \lexemeA{up}{RP}\&
          {$\vdots$}\&
            {$\vdots$}\&
                {$\vdots$}\&
                    {$\vdots$}\&
                      \lexemeA{all}{DT}\&
                              \lexemeA{of}{IN}\&
                                \lexemeA{the}{DT}\&
                                  \lexemeA{laundry}{NNS}\\
(h)\&
    \lexemeA{he}{PRP}\&
      \lexemeA{folded}{VBD}\&
        {$\vdots$}\&
          \lexemeA{under}{IN}\&
            \lexemeA{pressure}{NN}\&
                {$\vdots$}\&
                    {$\vdots$}\&
                      {$\vdots$}\& 
                              {$\vdots$}\&
                                {$\vdots$}\& 
                                  {$\vdots$}\\
                                    
  \&
    \&
      \&
        \&
          \&
            \&
                \&
                    \&
                      \&
                              \&
                                \&
                                  \&\\[2pt]
\end{deptext}
 \deproot[edge unit distance=3.2ex]{3}{anchor}
 \depedge[edge unit distance=2.5ex]{3}{2}{\ctA{nsubj}}
 \depedge[edge unit distance=2.5ex]{3}{4}{\ctA{prp}}
 \depedge[edge unit distance=2.2ex]{3}{6}{\ctA{nmod}}
 \depedge[edge unit distance=2.5ex]{6}{5}{\ctA{case}}
 \depedge[edge unit distance=1.5ex]{3}{9}{\ctA{dobj}}
 \depedge[edge unit distance=2.7ex]{9}{7}{\ctA{det}}
 \depedge[edge unit distance=2.5ex]{9}{8}{\ctA{amod}}
 \depedge[edge unit distance=2.7ex]{9}{12}{\ctA{nmod}}
 \depedge[edge unit distance=2.5ex]{12}{11}{\ctA{det}}
 \depedge[edge unit distance=2.7ex]{12}{10}{\ctA{case}}
\end{dependency} 
\caption{The distributional lexicon produced from the trees in Figure~\ref{fig:deptrees} with the elementary {\APT} for \lexeme{dry}{JJ} at the top, the elementary {\APT} for \lexeme{clothes}{NNS} in the middle, and the elementary {\APT} for \lexeme{folded}{VBD} at the bottom. Part of speech tags and weights have been omitted.}
\label{fig:lexicon}
\end{figure}

Figure~\ref{fig:lexicon} shows  three elementary {\APT}s that can be produced from the corpus shown in Figure~\ref{fig:deptrees}. On the far left we give the letter corresponding to the sentence in Figure~\ref{fig:deptrees} that generated the typed co-occurrences. Each column corresponds to one node in the {\APT}, giving the multiset of lexemes at that node. Weights are not shown, and only non-empty nodes are displayed. 

It is worth dwelling on the contents of the anchor node of the top {\APT} in Figure~\ref{fig:lexicon}, which is the elementary {\APT} for \lexeme{dry}{JJ}. The weighted multiset at  the anchor node is denoted $\apt{w}(\epsilon)$. The lexeme \lexeme{dry}{JJ} occurs three times, and the weight  $\apt{w}(\epsilon,\lexeme{dry}{JJ})$  reflects this count. Three other lexemes also occur at this same node: \lexeme{fizzy}{JJ}, \lexeme{white}{JJ} and \lexeme{clean}{JJ}. These lexemes arose from the following co-occurrences in trees in Figure~\ref{fig:deptrees}:  $\tco{\lexeme{dry}{JJ}}{\lexeme{fizzy}{JJ}}{\ctB{\inv{amod}}{amod}}$, $\tco{\lexeme{dry}{JJ}}{\lexeme{white}{JJ}}{\ctB{\inv{amod}}{amod}}$ and $\tco{\lexeme{dry}{JJ}}{\lexeme{clean}{JJ}}{\ctB{\inv{amod}}{amod}}$, all of which involve the co-occurrence type $\ctB{\inv{amod}}{amod}$. These lexemes appear in the  multiset $\apt{w}(\epsilon)$ because $\reduced{\ctB{\inv{amod}}{amod}}=\epsilon$.

\section{{\APT} Similarity}
\label{sec:similarity}

One of the most fundamental aspects of any treatment of distributional semantics is that it supports a way of measuring distributional similarity. In this section, we describe a straightforward way in which the similarity of two {\APT}s can be measured through a mapping from {\APT}s to vectors. 

First define the set of distributional features
\begin{equation}
	\features{C} = 
	\setof{\feat{\tau}{w'}}
	{$w'\in V$, $\tau\in\xoverline{R}^*R^*$ and $\weight{w}{\tau}{w'}>0$ for some $w\in V$}
\label{eq:features}
\end{equation}
The vector space that we use to encode {\APT}s  includes one dimension for each element of $\features{C}$, and we  use the pair $\feat{\tau}{w}$ to refer to its corresponding dimension. 

Given an {\APT} $\aptvar$, we denote the vectorized representation of $\aptvar$ with $\vectorise{\aptvar}$, and the value that the vector $\vectorise{\aptvar}$ has on dimension $\feat{\tau}{w'}$  is denoted $\dimension{\vectorise{\aptvar}}{\feat{\tau}{w'}}$.  For each $\feat{\tau}{w'}\in\features{C}$:
\begin{equation}
	\dimension{\vectorise{\apt{w}}}{\feat{\tau}{w'}} =
		\PW{\tau}{w}\,\weight{w}{\tau}{w'}	
\label{eq:vectorise}
\end{equation} 
where $\PW{\tau}{w}$ is a path weighting function which is intended to reflect the fact that not all of the distributional features are equally important in determining the distributional similarity of two {\APT}s.  Generally speaking, syntactically distant co-occurrences provide a weaker characterization of the semantics of a lexeme than co-occurrences that are syntactically closer. By multiplying each $\weight{w}{\tau}{w'}$ by $\PW{\tau}{w}$ we are able to capture this give a suitable instantiation of $\PW{\tau}{w}$.

One option for $\PW{\tau}{w}$ is to use $\pcond{\tau}{w}$, i.e. the probability that when randomly selecting one of the co-occurrences $\tco{w}{w'}{\tau'}$, where $w'$ can be any lexeme in $V$, $\tau'$ is the co-occurrence type $\tau$. We can estimate these path probabilities from the co-occurrence counts in $C$ as follows:
\begin{equation}
	\pcond{\tau}{w} = %
	\frac{%
		\cooccurs{\tco{w}{*}{\tau}}}%
		{\cooccurs{\tco{w}{*}{*}}}
\label{eq:prob4}
\end{equation}
where 
\[\begin{array}{ll@{\hspace*{6pt}}l}
	\cooccurs{\tco{w}{*}{\tau}} &=& 
		\sum_{w'\in V}\cooccurs{\tco{w}{w'}{\tau}}\\[2pt]
	\cooccurs{\tco{w}{*}{*}} &=& 
		\sum_{w'\in V}\sum_{\tau\in\bar{R}^*R^*}\cooccurs{\tco{w}{w'}{\tau}}
\end{array}\]
$\pcond{\tau}{w}$ typically falls off rapidly as a function of the length of $\tau$ as desired. 

The similarity of two {\APT}s, $\aptvar_1$ and $\aptvar_2$, which we  denote $\similar{\aptvar_1}{\aptvar_2}$, can be measured in terms of the similarity of  vectors $\vectorise{\aptvar_1}$ and $\vectorise{\aptvar_2}$. The similarity of vectors can be measured in a variety of ways~\cite{Lin_1998b,Lee_1999,Weeds_2005,Curran_2004}. One popular option involves the use of the cosine measure:
\begin{equation}
\similar{\aptvar_1}{\aptvar_2}= \cos(\vectorise{\aptvar_1},\vectorise{\aptvar_2})
\label{eq:sim}	
\end{equation}
It is common to apply cosine to vectors containing positive pointwise mutual information (PPMI) values.  If the weights used in the {\APT}s are counts or probabilities then they can be transformed into PPMI values at this point.

As a consequence of the fact that the different co-occurrence types of the co-occurrences associated with a lexeme are being differentiated, vectorized {\APT}s are much sparser than traditional vector representations used to model distributional semantics. This can be mitigated in various ways, including:
\begin{itemize}
\item reducing the granularity of the dependency relations and/or the part-of-speech tag-set;
\item applying various normalizations of lexemes such as case normalization, lemmatization, or stemming;
\item disregarding all distributional features involving co-occurrence types over a certain length;
\item applying some form of distributional smoothing, where  distributional features of a lexeme are inferred based on the features of distributionally similar lexemes.
\end{itemize}

\section{Distributional Composition}
\label{sec:composition}

In this section we turn to the central topic of the paper, namely distributional composition.  We begin with an informal explanation of our approach, and then present a more precise formalisation.

\subsection{Discussion of Approach}
\label{sec:approach}

Our starting point is the observation that although we have shown that all of the elementary {\APT}s in the distributional lexicon can be placed in the same vector space (see Section~\ref{sec:similarity}), there is an important sense in which {\APT}s for different parts of speech are not comparable.  For example, many of the dimensions that make sense for verbs, such as those involving a co-occurrence type  that begins with $\ctA{dobj}$ or $\ctA{nsubj}$, do not make sense for a noun. However, as we now explain, the co-occurrence type structure present in an {\APT} allows us to address this, making way for our definition of distributional composition.

Consider the {\APT}  for the lexeme \lexeme{dry}{JJ} shown at the top of Figure~\ref{fig:lexicon}. The anchor of this {\APT} is the node at which the lexeme \lexeme{dry}{JJ} appears. We can, however, take a different perspective on this {\APT}, for example, one in which the anchor is the node at which the lexemes \lexeme{bought}{VBD} and \lexeme{folded}{VBD} appear. This {\APT} is shown at the top of Figure~\ref{fig:dryoffset}. Adjusting the position of the anchor is significant because the starting point of the paths given by the co-occurrence types changes. For example, when the {\APT} shown at the top of Figure~\ref{fig:dryoffset} is applied to the co-occurrence type $\ctA{\inv{nsubj}}$, we reach the node at which the lexemes \lexeme{we}{PRP} and \lexeme{he}{PRP} appear. Thus, this {\APT}  can be seen as a characterisation of the distributional properties of the verbs that nouns that \lexeme{dry}{JJ}  modifies can take as their direct object. In fact, it looks rather like the elementary {\APT} for some verb. The lower tree in Figure~\ref{fig:dryoffset} shows the elementary {\APT} for \lexeme{clothes}{NNS} (the centre {\APT} shown in Figure~\ref{fig:lexicon}) where the anchor has been moved to the node at which the lexemes \lexeme{folded}{VBD}, \lexeme{hung}{VBD} and \lexeme{bought}{VBD} appear.

Notice that in both of the {\APT}s shown in Figure~\ref{fig:dryoffset} parts of the tree are shown in faded text. These are nodes and edges that are removed from the {\APT} as a result of where the anchor has been moved. The elementary tree for \lexeme{dry}{JJ} shown in Figure~\ref{fig:lexicon} reflects the fact that at least some of the nouns that \lexeme{dry}{JJ} modifies can be the direct object of a verb, or the subject of a verb. When we move the anchor, as shown at the top of Figure~\ref{fig:dryoffset}, we resolve this ambiguity to the case where the noun being modified is a direct object. The incompatible parts of the {\APT} are removed. This corresponds to restricting the co-occurrence types of composed {\APT}s to those that belong to the set $\xoverline{R}^*R^*$, just as was the case for elementary {\APT}s. For example, note that in the upper {\APT} of Figure~\ref{fig:dryoffset}, neither the path $\ctB{dobj}{\inv{nsubj}}$ from the node labeled with \lexeme{bought}{VBD} and \lexeme{folded}{VBD} to the node labeled \lexeme{caused}{VBD}, or the path $\ctC{dobj}{\inv{subj}}{\inv{dobj}}$ from the node labeled with \lexeme{bought}{VBD} and \lexeme{folded}{VBD} to the node labeled \lexeme{laughter}{NN} are in $\xoverline{R}^*R^*$.

\begin{figure}
\centering
\begin{dependency}
\begin{deptext}[column sep=.0cm, row sep=.0ex]
(a)\&
  \lexemeA{we}{PRP}\&
    \lexemeA{bought}{VBD}\& 
      {$\vdots$}\&
        \lexemeA{the}{DT}\&
          \lexemeA{slightly}{RB}\&
            \lexemeA{fizzy}{JJ}\&
              \lexemeA{wine}{NN}\&
                \textcolor{lightgray}{$\vdots$}\&
                  \textcolor{lightgray}{$\vdots$}
                  \\
\&
  {$\vdots$}\&
    {$\vdots$}\&
      {$\vdots$}\&
        {$\vdots$}\&
          {$\vdots$}\&
            \lexemeA{dry}{JJ}\&
              {$\vdots$}\&
                \textcolor{lightgray}{$\vdots$}\&
                  \textcolor{lightgray}{$\vdots$}
                  \\
\&
  {$\vdots$}\&
    {$\vdots$}\&
      {$\vdots$}\&
        {$\vdots$}\&
          {$\vdots$}\&
            \lexemeA{white}{JJ}\&
              {$\vdots$}\&
                \textcolor{lightgray}{$\vdots$}\&
                  \textcolor{lightgray}{$\vdots$}
                  \\
(b)\&
  {$\vdots$}\&
    {$\vdots$}\&
      \lexemeA{your}{PRP\$}\&
        {$\vdots$}\&
          {$\vdots$}\&
            \lexemeA{dry}{JJ}\&
              \lexemeA{joke}{NN}\&
                \textcolor{lightgray}{\lexemeA{caused}{VBD}}\&
                  \textcolor{lightgray}{\lexemeA{laughter}{NN}}
                  \\
(c)\&
  \lexemeA{he}{PRP}\&
    \lexemeA{folded}{VBD}\&
      {$\vdots$}\&
        \lexemeA{the}{DT}\&
          {$\vdots$}\&
            \lexemeA{clean}{JJ}\&
              \lexemeA{clothes}{NNS}\&
                \textcolor{lightgray}{$\vdots$}\&
                  \textcolor{lightgray}{$\vdots$}
                  \\
\&
  {$\vdots$}\&
    {$\vdots$}\&
      {$\vdots$}\&
        {$\vdots$}\&
          {$\vdots$}\&
            \lexemeA{dry}{JJ}\&
              {$\vdots$}\&
                \textcolor{lightgray}{$\vdots$}\&
                  \textcolor{lightgray}{$\vdots$}
                  \\          
  \&
    \&
      \&
        \&
          \&
            \&
                \&
                    \&
                      \&
                              \&
                                \&
                                  \&
                                  \\[4pt]
\end{deptext}
\deproot[edge unit distance=3.5ex]{3}{anchor}
\depedge[edge unit distance=2.5ex]{3}{2}{\ctA{nsubj}}
\depedge[edge unit distance=2ex]{3}{8}{\ctA{dobj}}
\depedge[edge unit distance=2ex]{8}{4}{\ctA{poss}}
\depedge[edge unit distance=2ex]{8}{5}{\ctA{det}}
\depedge[edge unit distance=2.5ex]{7}{6}{\ctA{advmod}}
\depedge[edge unit distance=2.5ex]{8}{7}{\ctA{amod}}
\depedge[edge unit distance=2.5ex, color=lightgray, label style={draw=lightgray}]{9}{8}{\textcolor{lightgray}{\ctA{nsubj}}}
\depedge[edge unit distance=2.5ex, color=lightgray, label style={draw=lightgray}]{9}{10}{\textcolor{lightgray}{\ctA{dobj}}}
\end{dependency}
\begin{dependency}
\begin{deptext}[column sep=.0cm, row sep=.0ex]
(c)\&
  {$\vdots$}\&
    \lexemeA{he}{PRP}\&
      \lexemeA{folded}{VDB}\&
        {$\vdots$}\&
              {$\vdots$}\&
                \lexemeA{the}{DT}\&
                  {$\vdots$}\&
                    \lexemeA{clean}{JJ}\&
                      \lexemeA{clothes}{NNS}\&
                        \textcolor{lightgray}{$\vdots$}\&
                            \textcolor{lightgray}{$\vdots$}\&
                                    {$\vdots$}\&
                                    \\
\&
  {$\vdots$}\&
    {$\vdots$}\&
      {$\vdots$}\&
        {$\vdots$}\&
              {$\vdots$}\&
                {$\vdots$}\&
                  {$\vdots$}\&
                    \lexemeA{dry}{JJ}\&
                      {$\vdots$}\&
                        \textcolor{lightgray}{$\vdots$}\&
                            \textcolor{lightgray}{$\vdots$}\&
                                    {$\vdots$}\&
                                    \\
(d)\&
  {$\vdots$}\&
    {$\vdots$}\&
      {$\vdots$}\&
        {$\vdots$}\&
              \lexemeA{your}{PRP\$}\& 
                {$\vdots$}\&
                  {$\vdots$}\&
                    {$\vdots$}\&
                      \lexemeA{clothes}{NNS}\&
                        \textcolor{lightgray}{\lexemeA{look}{VBP}}\&
                            \textcolor{lightgray}{\lexemeA{great}{JJ}}\&
                                    {$\vdots$}
                                    \\  
(e)\&
  \lexemeA{the}{DT}\& 
    \lexemeA{man}{NN}\&
      \lexemeA{hung}{VBD}\&
        \lexemeA{up}{RP}\&
              {$\vdots$}\&
                \lexemeA{the}{DT}\&
                  {$\vdots$}\&
                    \lexemeA{wet}{JJ}\&
                      \lexemeA{clothes}{NNS}\&
                        \textcolor{lightgray}{$\vdots$}\&
                            \textcolor{lightgray}{$\vdots$}\&
                                    {$\vdots$}
                                    \\  
(f)\&
  \lexemeA{a}{DT}\&
    \lexemeA{boy}{NN}\&
      \lexemeA{bought}{VBD}\&
        {$\vdots$}\&
              {$\vdots$}\&
                \lexemeA{some}{DT}\&
                  \lexemeA{very}{RB}\&
                    \lexemeA{expensive}{JJ}\&
                      \lexemeA{clothes}{NNS}\&
                        \textcolor{lightgray}{$\vdots$}\&
                            \textcolor{lightgray}{$\vdots$}\&
                                    \lexemeA{yesterday}{NN}
                                    \\
  \&
    \&
      \&
        \&
          \&
            \&
                \&
                    \&
                      \&
                              \&
                                \&
                                  \&
                                  \\[4pt]
\end{deptext} 
 \deproot[edge unit distance=4.8ex]{4}{anchor}
 \depedge[edge unit distance=2.2ex]{3}{2}{\ctA{det}}
 \depedge[edge unit distance=2.5ex]{4}{3}{\ctA{nsubj}}
 \depedge[edge unit distance=2.5ex]{4}{5}{\ctA{prp}}
 \depedge[edge unit distance=2.3ex]{10}{6}{\ctA{poss}}
 \depedge[edge unit distance=2.2ex]{10}{7}{\ctA{det}}
 \depedge[edge unit distance=2.5ex]{9}{8}{\ctA{advmod}}
 \depedge[edge unit distance=2.5ex]{10}{9}{\ctA{amod}}
 \depedge[edge unit distance=2.5ex, color=lightgray, label style={draw=lightgray}]{11}{10}{\textcolor{lightgray}{\ctA{nsubj}}}
 \depedge[edge unit distance=2.1ex]{4}{10}{\ctA{dobj}}
 \depedge[edge unit distance=2.7ex, color=lightgray, label style={draw=lightgray}]{11}{12}{\textcolor{lightgray}{\ctA{xcomp}}}
 \depedge[edge unit distance=1.6ex]{4}{13}{\ctA{tmod}}
\end{dependency} 
\caption{The elementary {\APT}s for \protect\lexeme{dry}{JJ} and \protect\lexeme{clothes}{NNS} with anchors offset. }
\label{fig:dryoffset}
\end{figure}

Given a sufficiently rich elementary {\APT} for \lexeme{dry}{JJ}, those verbs that have nouns that \lexeme{dry}{JJ} can plausibly modify as direct objects have elementary {\APT}s that are in some sense ``compatible'' with the {\APT} produced by shifting the anchor node as illustrated at the top of Figure~\ref{fig:dryoffset}. An example is the {\APT} for \lexeme{folded}{VBD} shown at the bottom of Figure~\ref{fig:lexicon}. Loosely speaking, this means that when applied to the same co-occurrence type, the  {\APT} in Figure~\ref{fig:dryoffset} and the {\APT} at the bottom of Figure~\ref{fig:lexicon} are generally expected to give sets of lexemes with related elements. 

By moving the anchors of the {\APT} for \lexeme{dry}{JJ} and \lexeme{clothes}{NNS} as in Figure~\ref{fig:dryoffset}, we have, in effect, aligned all of the nodes of the {\APT}s for \lexeme{dry}{JJ} and \lexeme{clothes}{NN} with the nodes they correspond to in the {\APT} for \lexeme{folded}{VBD}.  Not only does this make it possible, in principle at least, to establish whether or not the composition of \lexeme{dry}{JJ}, \lexeme{clothes}{NNS} and \lexeme{folded}{VBD} is plausible, it provides the basis for the contextualization of {\APT}s, as we now explain.

Recall that elementary {\APT}s are produced by aggregating contexts taken from all of the occurrences of the lexeme in a corpus. As described in the introduction, we need a way to contextualize aggregated {\APT}s in order to produce a fine-grained characterization of the distributional semantics of the lexeme in context. There are two distinct aspects to the contextualization of {\APT}s, both of which can be captured through {\APT} composition: \textbf{co-occurrence filtering} --- the down-weighting of co-occurrences that are not  compatible with the way the lexeme is being used in its current context; and \textbf{co-occurrence embellishment} --- the up-weighting of compatible co-occurrences that appear in the {\APT}s for the lexemes with which it is being composed.

Both co-occurrence filtering and co-occurrence embellishment can be achieved through {\APT} composition. The process of composing the elementary {\APT}s for the lexemes that appear in a phrase involves two distinct steps. First, the elementary {\APT}s for each of the lexemes being composed are aligned in a way that is determined by the dependency tree for the phrase. The result of  this alignment of the elementary {\APT}s, is that each node in one of the {\APT}s is matched up with (at most) one of the nodes in each of the other {\APT}s. The second step of this process involves merging nodes that have been matched up with one another in order to produce the resulting composed {\APT} that represents the distributional semantics of the dependency tree. It is during this second step that we are in a position to determine those co-occurrences that are compatible across the nodes that have been matched up.

\begin{figure}
	\centering
\includegraphics[scale=0.4]{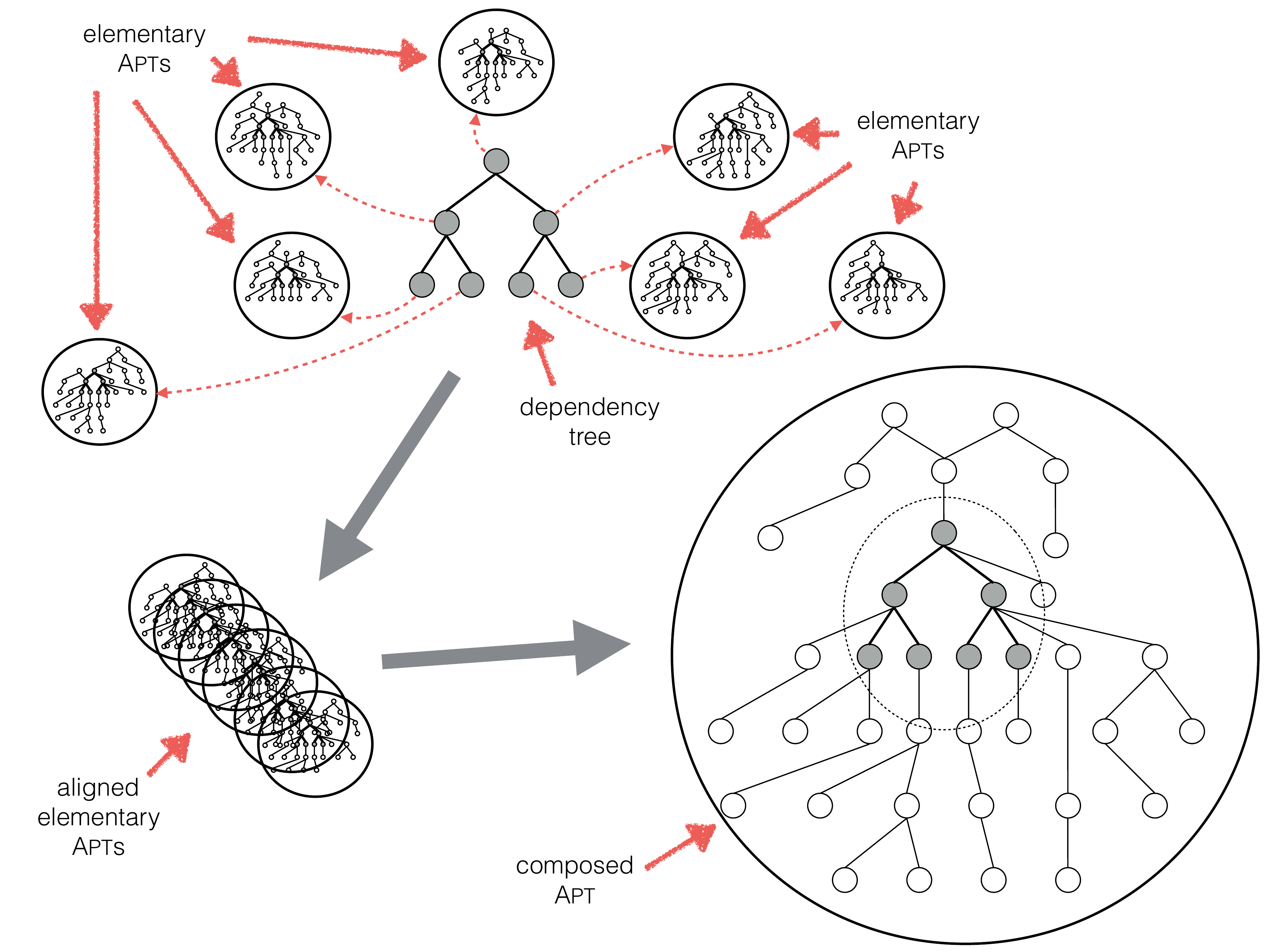}
\caption{Composition of {\APT}s.}
\label{fig:treecontext}
\end{figure}

Figure~\ref{fig:treecontext} illustrates the composition of  {\APT}s on the basis of a dependency tree shown in the upper centre of the figure. In the lower right, the figure shows the full {\APT} that  results from merging the six aligned {\APT}s, one for each of the lexemes in the dependency tree. Each node in the dependency tree is labeled with a lexeme, and around the dependency tree, we show the elementary {\APT}s  for each lexeme.  The six elementary {\APT}s are aligned on the basis of the position of their lexeme in the dependency tree. Note that the tree  shown in grey within the {\APT} is structurally identical to the dependency tree in the upper centre of the figure. The nodes of the dependency tree are labeled with single lexemes, whereas each node of the {\APT} is labeled by a weighted lexeme multiset. The lexeme labelling a node in the dependency tree is one of the lexemes found in the weighted lexeme multiset associated with the corresponding node within the {\APT}. We refer to the nodes in the composed {\APT} that come from nodes in the dependency tree (the grey nodes) as the \textbf{internal context}, and the remaining nodes as the \textbf{external context}.

\begin{figure}
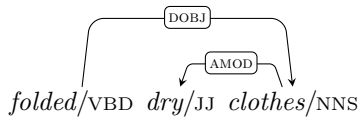

\centering
\begin{dependency}
\begin{deptext}[column sep=.0cm, row sep=.1ex]
\lexeme{folded}{VBD} \& \lexeme{dry}{JJ} \& \lexeme{clothes}{NNS} \\
   \end{deptext}
   \depedge[edge unit distance=3ex]{1}{3}{\ctA{dobj}}
   \depedge[edge unit distance=2ex]{3}{2}{\ctA{amod}}
\end{dependency}\\[8pt]
\caption{A dependency tree that generates the alignment shown in Figure~\ref{fig:aligned}.}
\label{fig:depexample}
\end{figure}

\begin{center}
\begin{sidewaysfigure}
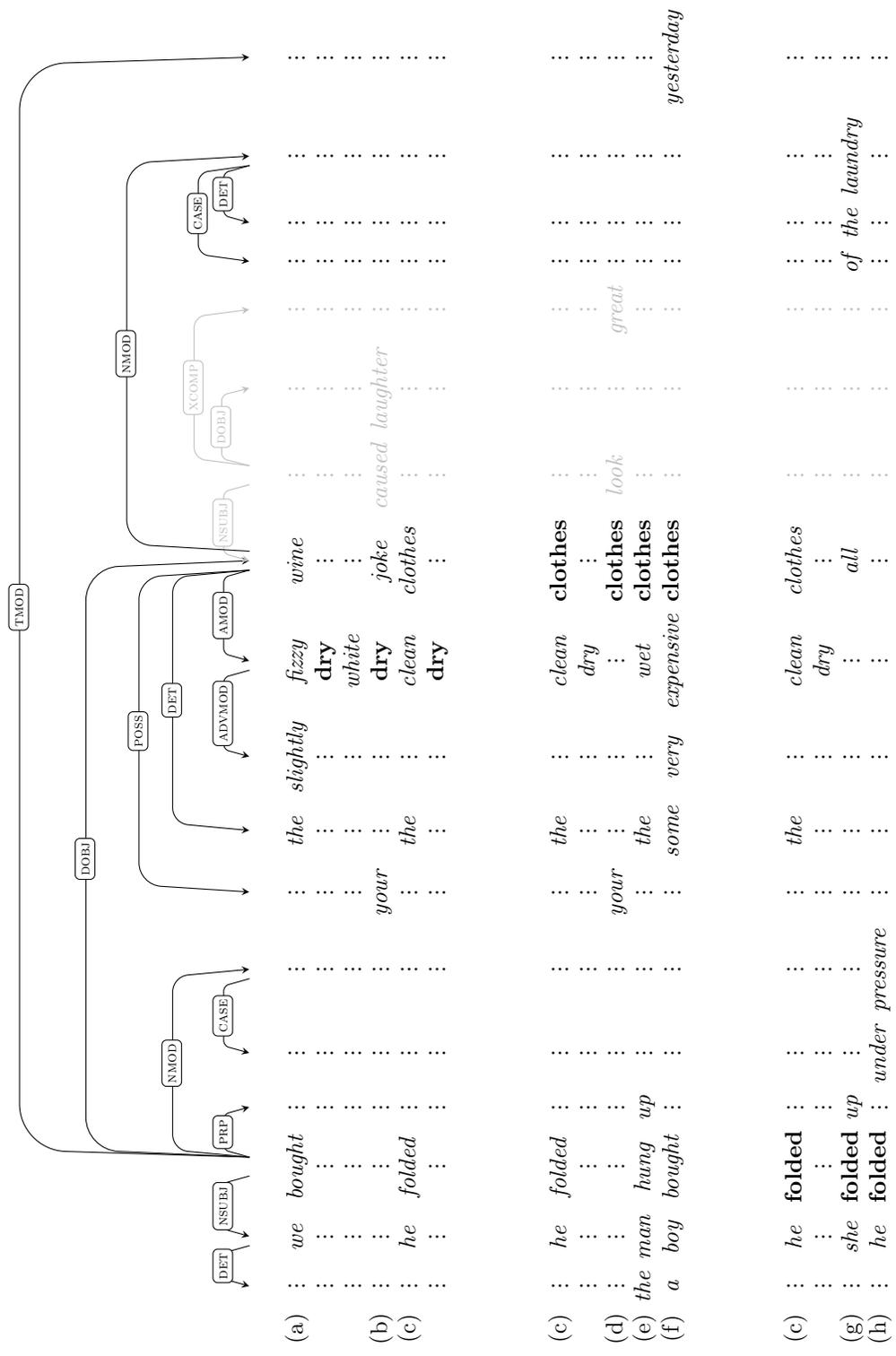

\begin{dependency}
\begin{deptext}[column sep=.0cm, row sep=.0ex] 
\&
  \&
    \&
      \&
        \&
          \&
            \&
              \&
                \&
                  \&
                    \&
                      \&
                        \&
                          \&
                            \&
                              \&
                                \&
                                  \&
                                    \&\\[2pt]
(a)\&
  {$\vdots$}\&
    \lexemeA{we}{PRP}\& 
      \lexemeA{bought}{VBD}\&
        {$\vdots$}\&
          {$\vdots$}\&
            {$\vdots$}\&
              {$\vdots$}\&
                \lexemeA{the}{DT}\&
                  \lexemeA{slightly}{RB}\&
                    \lexemeA{fizzy}{JJ}\&
                      \lexemeA{wine}{NN}\&
                        \textcolor{lightgray}{$\vdots$}\&
                          \textcolor{lightgray}{$\vdots$}\&
                            \textcolor{lightgray}{$\vdots$}\&
                              {$\vdots$}\&
                                {$\vdots$}\&
                                  {$\vdots$}\&
                                    {$\vdots$}\\  
\&
  {$\vdots$}\&
    {$\vdots$}\&
      {$\vdots$}\&
        {$\vdots$}\&
          {$\vdots$}\&
            {$\vdots$}\&
              {$\vdots$}\&
                {$\vdots$}\&
                  {$\vdots$}\&
                    \lexemeA{\bf dry}{\bf JJ}\&
                      {$\vdots$}\&
                        \textcolor{lightgray}{$\vdots$}\&
                          \textcolor{lightgray}{$\vdots$}\&
                            \textcolor{lightgray}{$\vdots$}\&
                              {$\vdots$}\&
                                {$\vdots$}\&
                                  {$\vdots$}\&
                                    {$\vdots$}\\  
\&
  {$\vdots$}\&
    {$\vdots$}\&
      {$\vdots$}\&
        {$\vdots$}\&
          {$\vdots$}\&
            {$\vdots$}\&
              {$\vdots$}\&
                {$\vdots$}\&
                  {$\vdots$}\&
                    \lexemeA{white}{JJ}\&
                      {$\vdots$}\&
                        \textcolor{lightgray}{$\vdots$}\&
                          \textcolor{lightgray}{$\vdots$}\&
                            \textcolor{lightgray}{$\vdots$}\&
                              {$\vdots$}\&
                                {$\vdots$}\&
                                  {$\vdots$}\&
                                    {$\vdots$}\\
                                  
(b)\&
  {$\vdots$}\&
    {$\vdots$}\&
      {$\vdots$}\&
        {$\vdots$}\&
          {$\vdots$}\&
            {$\vdots$}\&
              \lexemeA{your}{PRP\$}\&
                {$\vdots$}\&
                  {$\vdots$}\&
                    \lexemeA{\bf dry}{\bf JJ}\&
                      \lexemeA{joke}{NN}\&
                        \textcolor{lightgray}{\lexemeA{caused}{VBD}}\&
                          \textcolor{lightgray}{\lexemeA{laughter}{NN}}\&
                            \textcolor{lightgray}{$\vdots$}\&
                              {$\vdots$}\&
                                {$\vdots$}\&
                                  {$\vdots$}\&
                                    {$\vdots$}\\
(c)\&
  {$\vdots$}\&
    \lexemeA{he}{PRP}\&
      \lexemeA{folded}{VDB}\&
        {$\vdots$}\&
          {$\vdots$}\&
            {$\vdots$}\&
              {$\vdots$}\&
                \lexemeA{the}{DT}\&
                  {$\vdots$}\&
                    \lexemeA{clean}{JJ}\&
                      \lexemeA{clothes}{NNS}\&
                        \textcolor{lightgray}{$\vdots$}\&
                          \textcolor{lightgray}{$\vdots$}\&
                            \textcolor{lightgray}{$\vdots$}\&
                              {$\vdots$}\&
                                {$\vdots$}\&
                                  {$\vdots$}\&
                                    {$\vdots$}\\
\&
  {$\vdots$}\&
    {$\vdots$}\&
      {$\vdots$}\&
        {$\vdots$}\&
          {$\vdots$}\&
            {$\vdots$}\&
              {$\vdots$}\&
                {$\vdots$}\&
                  {$\vdots$}\&
                    \lexemeA{\bf dry}{\bf JJ}\&
                      {$\vdots$}\&
                        \textcolor{lightgray}{$\vdots$}\&
                          \textcolor{lightgray}{$\vdots$}\&
                            \textcolor{lightgray}{$\vdots$}\&
                              {$\vdots$}\&
                                {$\vdots$}\&
                                  {$\vdots$}\&
                                    {$\vdots$}\&\\[40pt]
(c)\&
  {$\vdots$}\&
    \lexemeA{he}{PRP}\&
      \lexemeA{folded}{VDB}\&
        {$\vdots$}\&
          {$\vdots$}\&
            {$\vdots$}\&
              {$\vdots$}\&
                \lexemeA{the}{DT}\&
                  {$\vdots$}\&
                    \lexemeA{clean}{JJ}\&
                      \lexemeA{\bf clothes}{\bf NNS}\&
                        \textcolor{lightgray}{$\vdots$}\&
                          \textcolor{lightgray}{$\vdots$}\&
                            \textcolor{lightgray}{$\vdots$}\&
                              {$\vdots$}\&
                                {$\vdots$}\&
                                  {$\vdots$}\&
                                    {$\vdots$}\&\\
\&
  {$\vdots$}\&
    {$\vdots$}\&
      {$\vdots$}\&
        {$\vdots$}\&
          {$\vdots$}\&
            {$\vdots$}\&
              {$\vdots$}\&
                {$\vdots$}\&
                  {$\vdots$}\&
                    \lexemeA{dry}{JJ}\&
                      {$\vdots$}\&
                        \textcolor{lightgray}{$\vdots$}\&
                          \textcolor{lightgray}{$\vdots$}\&
                            \textcolor{lightgray}{$\vdots$}\&
                              {$\vdots$}\&
                                {$\vdots$}\&
                                  {$\vdots$}\&
                                    {$\vdots$}\&\\
(d)\&
  {$\vdots$}\&
    {$\vdots$}\&
      {$\vdots$}\&
        {$\vdots$}\&
          {$\vdots$}\&
            {$\vdots$}\&
              \lexemeA{your}{PRP\$}\& 
                {$\vdots$}\&
                  {$\vdots$}\&
                    {$\vdots$}\&
                      \lexemeA{\bf clothes}{\bf NNS}\&
                        \textcolor{lightgray}{\lexemeA{look}{VBP}}\&
                          \textcolor{lightgray}{$\vdots$}\&
                            \textcolor{lightgray}{\lexemeA{great}{JJ}}\&
                              {$\vdots$}\&
                                {$\vdots$}\&
                                  {$\vdots$}\&
                                    {$\vdots$}\\  
(e)\&
  \lexemeA{the}{DT}\& 
    \lexemeA{man}{NN}\&
      \lexemeA{hung}{VBD}\&
        \lexemeA{up}{RP}\&
          {$\vdots$}\&
            {$\vdots$}\&
              {$\vdots$}\&
                \lexemeA{the}{DT}\&
                  {$\vdots$}\&
                    \lexemeA{wet}{JJ}\&
                      \lexemeA{\bf clothes}{\bf NNS}\&
                        \textcolor{lightgray}{$\vdots$}\&
                          \textcolor{lightgray}{$\vdots$}\&
                            \textcolor{lightgray}{$\vdots$}\&
                              {$\vdots$}\&
                                {$\vdots$}\&
                                  {$\vdots$}\&
                                    {$\vdots$}\\  
(f)\&
  \lexemeA{a}{DT}\&
    \lexemeA{boy}{NN}\&
      \lexemeA{bought}{VBD}\&
        {$\vdots$}\&
          {$\vdots$}\&
            {$\vdots$}\&
              {$\vdots$}\&
                \lexemeA{some}{DT}\&
                  \lexemeA{very}{RB}\&
                    \lexemeA{expensive}{JJ}\&
                      \lexemeA{\bf clothes}{\bf NNS}\&
                        \textcolor{lightgray}{$\vdots$}\&
                          \textcolor{lightgray}{$\vdots$}\&
                            \textcolor{lightgray}{$\vdots$}\&
                              {$\vdots$}\&
                                {$\vdots$}\&
                                  {$\vdots$}\&
                                    \lexemeA{yesterday}{NN}\\[40pt]
(c)\&
  {$\vdots$}\&
    \lexemeA{he}{PRP}\&
      \lexemeA{\bf folded}{\bf VDB}\&
        {$\vdots$}\&
          {$\vdots$}\&
            {$\vdots$}\&
              {$\vdots$}\&
                \lexemeA{the}{DT}\&
                  {$\vdots$}\&
                    \lexemeA{clean}{JJ}\&
                      \lexemeA{clothes}{NNS}\&
                        \textcolor{lightgray}{$\vdots$}\&
                          \textcolor{lightgray}{$\vdots$}\&
                            \textcolor{lightgray}{$\vdots$}\&
                              {$\vdots$}\&
                                {$\vdots$}\&
                                  {$\vdots$}\&
                                    {$\vdots$}\\
\&
  {$\vdots$}\&
    {$\vdots$}\&
      {$\vdots$}\&
        {$\vdots$}\&
          {$\vdots$}\&
            {$\vdots$}\&
              {$\vdots$}\&
                {$\vdots$}\&
                  {$\vdots$}\&
                    \lexemeA{dry}{JJ}\&
                      {$\vdots$}\&
                        \textcolor{lightgray}{$\vdots$}\&
                          \textcolor{lightgray}{$\vdots$}\&
                            \textcolor{lightgray}{$\vdots$}\&
                              {$\vdots$}\&
                                {$\vdots$}\&
                                  {$\vdots$}\&
                                    {$\vdots$}\&\\
(g)\&
  {$\vdots$}\&
    \lexemeA{she}{PRP}\&
      \lexemeA{\bf folded}{\bf VBD}\&
        \lexemeA{up}{RP}\&
          {$\vdots$}\&
            {$\vdots$}\&
              {$\vdots$}\&
                {$\vdots$}\&
                  {$\vdots$}\& 
                    {$\vdots$}\&
                      \lexemeA{all}{DT}\&
                        \textcolor{lightgray}{$\vdots$}\&
                          \textcolor{lightgray}{$\vdots$}\&
                            \textcolor{lightgray}{$\vdots$}\&
                              \lexemeA{of}{IN}\&
                                \lexemeA{the}{DT}\&
                                  \lexemeA{laundry}{NNS}\&
                                    {$\vdots$}\\  
(h)\&
  {$\vdots$}\& 
    \lexemeA{he}{PRP}\&
      \lexemeA{\bf folded}{\bf VBD}\&
        {$\vdots$}\&
          \lexemeA{under}{IN}\&
            \lexemeA{pressure}{NN}\&
              {$\vdots$}\&
                {$\vdots$}\&
                  {$\vdots$}\&
                    {$\vdots$}\&
                      {$\vdots$}\& 
                        \textcolor{lightgray}{$\vdots$}\&
                          \textcolor{lightgray}{$\vdots$}\& 
                            \textcolor{lightgray}{$\vdots$}\& 
                              {$\vdots$}\&
                                {$\vdots$}\& 
                                  {$\vdots$}\& 
                                    {$\vdots$}\\
                                    
\&
  \&
    \&
      \&
        \&
          \&
            \&
              \&
                \&
                  \&
                    \&
                      \&
                        \&
                          \&
                            \&
                              \&
                                \&
                                  \&
                                    \&\\[2pt]
\end{deptext}
 \depedge[edge unit distance=2.5ex]{3}{2}{\ctA{det}}
 \depedge[edge unit distance=2.5ex]{4}{3}{\ctA{nsubj}}
 \depedge[edge unit distance=2.5ex]{4}{5}{\ctA{prp}}
 \depedge[edge unit distance=2.0ex]{4}{12}{\ctA{dobj}}
 \depedge[edge unit distance=2.5ex]{12}{9}{\ctA{det}}
 \depedge[edge unit distance=2.7ex]{12}{8}{\ctA{poss}}
 \depedge[edge unit distance=2.5ex]{11}{10}{\ctA{advmod}}
 \depedge[edge unit distance=2.5ex]{12}{11}{\ctA{amod}}
 \depedge[edge unit distance=2.5ex, color=lightgray, label style={draw=lightgray}]{13}{12}{\textcolor{lightgray}{\ctA{nsubj}}}
 \depedge[edge unit distance=2.7ex, color=lightgray, label style={draw=lightgray}]{13}{14}{\textcolor{lightgray}{\ctA{dobj}}}
 \depedge[edge unit distance=2.7ex, color=lightgray, label style={draw=lightgray}]{13}{15}{\textcolor{lightgray}{\ctA{xcomp}}}
 \depedge[edge unit distance=1.5ex]{4}{19}{\ctA{tmod}}
 \depedge[edge unit distance=2.0ex]{12}{18}{\ctA{nmod}}
 \depedge[edge unit distance=2.5ex]{18}{17}{\ctA{det}}
 \depedge[edge unit distance=2.5ex]{18}{16}{\ctA{case}}
 \depedge[edge unit distance=2.5ex]{4}{7}{\ctA{nmod}}
 \depedge[edge unit distance=2.5ex]{7}{6}{\ctA{case}}
\end{dependency} 
\caption{Vertically aligned {\APT}s for \protect\lexeme{dry}{JJ}, \protect\lexeme{clothes}{NNS} and \lexeme{folded}{VBD} as determined by the tree in Figure~\ref{fig:depexample}. The letters in brackets on the left refer to the dependency trees shown in Figure~\ref{fig:deptrees} from which these {\APT}s are constructed. For space reasons, part of speech tags have been omitted. }
\label{fig:aligned}
\end{sidewaysfigure}
\end{center}

As we have seen, the alignment of {\APT}s can be achieved by adjusting the location of the anchor. The specific adjustments to the anchor locations are determined by the dependency tree for the phrase.  For example, Figure~\ref{fig:depexample} shows a dependency analysis of the phrase \emph{folded dry clothes}. To align the elementary {\APT}s for the lexemes in this tree, we do the following.
\begin{itemize}
\item 	The anchor of the elementary {\APT} for \lexeme{dry}{JJ} is moved to the node on which the \lexeme{bought}{VBD} and \lexeme{folded}{VBD} lie. This is the {\APT} shown at the top of Figure~\ref{fig:aligned}. This  change of anchor location is determined by the path from the \lexeme{dry}{JJ} to \lexeme{folded}{VBD} in the tree in Figure~\ref{fig:depexample}, i.e. $\ctB{\inv{amod}}{\inv{dobj}}$.
\item The anchor of the elementary {\APT} for \lexeme{clothes}{NNS} is moved to the node on which \lexeme{folded}{VBD}, \lexeme{hung}{VBD} and \lexeme{bought}{VBD} lie. This is the {\APT} shown at the bottom of Figure~\ref{fig:dryoffset}. This change of anchor location is determined by the path  from the \lexeme{clothes}{NNS} to \lexeme{folded}{VBD} in the tree in Figure~\ref{fig:depexample}, i.e. $\ctA{\inv{dobj}}$.

\item The anchor of the elementary {\APT} for \lexeme{folded}{VBD} has been left unchanged because there is an empty path from from the \lexeme{folded}{VBD} to \lexeme{folded}{VBD} in the tree in Figure~\ref{fig:depexample}.
\end{itemize}

Figure~\ref{fig:aligned} shows the three elementary {\APT}s for the lexemes \lexeme{dry}{JJ}, \lexeme{clothes}{NNS} and \lexeme{folded}{VPD} which  have been aligned as determined by the dependency tree shown in Figure~\ref{fig:depexample}. Each column of lexemes appear at nodes that have been aligned with one another. For example, in the third column from the left, we see that the following three nodes have been aligned: (i)~the node in the elementary {\APT} for \lexeme{dry}{JJ} at which \lexeme{bought}{VBD} and \lexeme{folded}{VBD} appear; (ii)~the node in the elementary {\APT} for \lexeme{clothes}{NNS} at which \lexeme{folded}{VBD}, \lexeme{hung}{VBD} and \lexeme{bought}{VBD} appear; and (iii)~the anchor node of the elementary {\APT} for \lexeme{folded}{VBD}, i.e the node at which \lexeme{folded}{VBD} appears. In the second phase of composition, these three nodes are merged together to produce a single node in the composed {\APT}. 

Before we discuss how the nodes in aligned {\APT}s are merged, we formalize the notion of {\APT} alignment. We do this by first defining so-called offset {\APT}s, which formalizes the idea of adjusting the location of an anchor. We then define how to align all of the {\APT}s for the lexemes in a phrase based on a dependency tree.

\subsection{Offset {\APT}s}
\label{sec:offset}

Given some offset, $\delta$, a string in $\xoverline{R}^*R^*$, the {\APT} $\aptvar$ when offset by $\delta$ is denoted $\aptvar^{\delta}$. Offsetting an {\APT} by $\delta$ involves moving the anchor to the position reached by following the path $\delta$ from the original anchor position. 
%
%
In order to define $\aptvar^{\delta}$, we must define $\aptvar^{\delta}(\tau,w')$ for each $\tau\in\xoverline{R}^*R^*$ and $w'\in V$, or in terms of our alternative tree-based representation, we need to specify the $\tau'$ such that $\aptvar^{\delta}(\tau)$ and $\aptvar(\tau')$ yield the same node (weighted lexeme multiset).

As shown in the Equation~\ref{eq:offset} below, path offset can  be specified  by making use of the co-occurrence type reduction operator that was introduced in Section~\ref{sec:APTs}. Given a string $\delta$ in $\xoverline{R}^*R^*$ and an {\APT} $\aptvar$,  the offset {\APT} $\aptvar^{\delta}$ is defined as follows. For each $\tau\in\xoverline{R}^*R^*$ and $w\in V$:
\begin{equation}
\aptvar^{\delta}(\tau,w)=\aptvar(\reduced{\delta\tau},w)
\label{eq:offset}
\end{equation}
or equivalently, for each $\tau\in\xoverline{R}^*R^*$:
\begin{equation}
\aptvar^{\delta}(\tau)=\aptvar(\reduced{\delta\tau})
\label{eq:offsetalt}
\end{equation}
As required, Equation~\ref{eq:offset} defines $\aptvar^{\delta}$ by specifying the weighted lexeme multiset we get when $\aptvar^{\delta}$ is applied to co-occurrence type $\tau$  as being the  lexeme multiset that $\aptvar$  produces when applied to the co-occurrence type $\reduced{\delta\tau}$.

As an illustrative example, consider the {\APT} shown at the top of Figure~\ref{fig:lexicon}. Let us call this {\APT} $\aptvar$. Note that $\aptvar$ is anchored at the node where the lexeme \lexeme{dry}{JJ} appears. Consider the {\APT} produced when we apply the offset $\ctB{\inv{amod}}{\inv{dobj}}$. This is shown at the top of Figure~\ref{fig:dryoffset}. Let us refer to this {\APT} as $\aptvar'$. The anchor of $\aptvar'$ is the node at which the lexemes \lexeme{bought}{VDB} and \lexeme{folded}{VBD} appear.  Now we show how the two nodes $\aptvar'(\ctA{nsubj})$ and $\aptvar'(\ctC{dobj}{amod}{advmod})$ are defined in terms of $\aptvar$ on the basis of Equation~\ref{eq:offsetalt}. In both cases the offset $\delta=\ctB{\inv{amod}}{\inv{dobj}}$. 
\begin{itemize}
\item 	
For the case where $\tau=\ctA{nsubj}$ we have 
\begin{equation*}
\begin{split}
\aptvar'(\ctA{nsubj})&=\aptvar(\reduced{\ctC{\inv{amod}}{\inv{dobj}}{nsubj}})\\
&=\aptvar(\ctC{\inv{amod}}{\inv{dobj}}{nsubj})
\end{split}
\end{equation*}
With respect to the anchor of $\aptvar$, this correctly addresses the node at which the lexemes \lexeme{we}{PRP} and \lexeme{he}{PRP} appear.
\item 
Where $\tau=\ctC{dobj}{amod}{advmod}$ we have
\begin{equation*}
\begin{split}
\aptvar'(\ctC{dobj}{amod}{advmod})
&=\aptvar(\reduced{\ctE{\inv{amod}}{\inv{dobj}}{dobj}{amod}{advmod}})\\
&=\aptvar(\reduced{\ctC{\inv{amod}}{amod}{advmod}})\\
&=\aptvar(\reduced{\ctA{advmod}})\\
&=\aptvar(\ctA{advmod})
\end{split}
\end{equation*}

With respect to the anchor of $\aptvar$, this correctly addresses the node at which the lexeme \lexeme{slightly}{RB} appears.
\end{itemize}

In practice, the offset {\APT} $\aptvar^{\delta}$ can be obtained by prepending the inverse of the path offset, $\delta^{-1}$, to all of the co-occurrence types in $\aptvar$ and then repeatedly applying the reduction operator until no further reductions are possible.  In other words, if $\tau$ addresses a node in $\aptvar$, then $\tau'$ addresses a node in $\aptvar^{\delta}$ iff 
$\tau' = \reduced{\delta^{-1}\tau}$ 
and $\tau'\in\xoverline{R}^*R^*$.

\subsection{Syntax-driven {\APT} Alignment}
\label{sec:alignment}

We now make use of offset {\APT}s, as defined in Equation~\ref{eq:offset}, as a way to align all of the {\APT}s associated with a dependency tree. Consider the following scenario:
\begin{itemize}
	\item $w_1\ldots w_n$ is a the phrase (or sentence)  where each $w_i\in V$ for $1\le i\le n$;
	\item $t\in T_{V,R}$ is a dependency analysis of the string $w_1\ldots w_n$;
	\item $w_h$ is the lexeme at the root of $t$. In other words, $h$ is the position (index) in the phrase at which the head appears;
	\item $\apt{w_i}$ is the elementary {\APT} for $w_i$ for each $i$, $1\le i\le n$; and
	\item  $\delta_i$, the offset of $w_i$ in $t$ with respect to the root, is the path in $t$ from $w_i$ to $w_h$. In other words, $\tco{w_i}{w_h}{\delta_i}$ is a co-occurrence in ${t}$ for each $i$, $1\le i\le n$. Note that $\delta_h=\epsilon$.
\end{itemize}

We define the distributional semantics for the tree $t$, denoted   $\comptree{t}$, as follows:
\begin{equation}
	\comptree{t}=%
		\merge{\set{\offapt{w_1}{\delta_1},\ldots,\offapt{w_n}{\delta_n}}}
\label{eq:composition} 
\end{equation}
The definition of $\merge$ is considered in Section~\ref{sec:merging}. In general, $\merge$ operates on a set of $n$ aligned {\APT}s, merging them into a single {\APT}. The multiset at each node in the resulting {\APT} is formed by merging $n$ multisets, one from each of the elements of $\set{\offapt{w_1}{\delta_1},\ldots,\offapt{w_n}{\delta_n}}$. It is this multiset merging operation that we focus on in Section~\ref{sec:merging}.

Although $\comptree{t}$ can be taken to be the distributional semantics of the tree as a whole, the same {\APT}, when associated with different anchors (i.e. when offset in some appropriate way) provides a representation of each of the contextualized lexemes that appear in the tree.

For each $i$, for $1\le i\le n$, the {\APT} for $w_i$ when contextualized by its role in the dependency tree $t$, denoted $\compapt{w_i}{t}$, is the {\APT} that satisfies the equality: 
\begin{equation}
\compapt{w_i}{t}^{\delta_i}=\comptree{t}	\label{eq:contextAPTa}
\end{equation}
Alternatively, this can also be expressed with the  equality:
\begin{equation}
\compapt{w_i}{t}=\comptree{t}^{\inverse{\delta_i}}
\label{eq:contextAPTb}
\end{equation}
Note that $\compapt{w_h}{t}$ and $\comptree{t}$ are identical. In other words, we take the representation of the distributional semantics of a dependency tree to be the {\APT} for the lexeme at the root of that tree that has been contextualized by the other lexemes appearing below it in the tree.

Equation~\ref{eq:composition} defined {\APT} composition as a ``one-step'' process in the sense that all of the $n$ elementary {\APT}s that are associated with nodes in the dependency tree are composed at once to produce the resulting (composed) {\APT}. There are, however, alternative strategies that could be formulated. One possibility is fully incremental left-to-right composition, where, working left-to-right through the string of lexemes, the elementary {\APT}s for the first two lexemes are composed, with the resulting {\APT} then being composed with the elementary {\APT} for the third lexeme, and so on. It is always possible to compose {\APT}s in this fully incremental way, whatever the structure in the dependency tree.  The tree structure, is however, critical in determining how the adjacent {\APT}s need to be aligned. 

\subsection{Merging Aligned {\APT}s}
\label{sec:merging}

We now turn to the question of how to implement the function $\merge$ which appears in Equation~\ref{eq:composition}. $\merge$ takes a set of $n$ aligned {\APT}s, $\set{\aptvar_1,\ldots\aptvar_n}$,  one for each node in the dependency tree $t$. It merges the {\APT}s together node by node to produce a single {\APT}, $\merge{\set{\aptvar_1,\ldots\aptvar_n}}$, that represents the semantics of the dependency tree. Our discussion, therefore,  addresses the question of how to merge the multisets that appear at nodes that are aligned with each other and form the nodes of the {\APT} being produced. 

The elementary {\APT} for a lexeme expresses those co-occurrences that are \textbf{distributionally compatible} with the lexeme given the corpus.  When lexemes in some phrase are composed, our objective is to capture the extent to which the co-occurrences arising  in the elementary {\APT}s are mutually compatible with the phrase as a whole. Once the elementary {\APT}s that are being composed have been aligned, we are in a position to determine the extent to which co-occurrences are mutually compatible: co-occurrences that need to be compatible with one another are brought together through the alignment. We  consider two alternative ways in which this can be achieved.

We begin with $\merge_{\MIN}$ which provides a tight implementation of the mutual compatibility of co-occurrences. In particular, a co-occurrence is only deemed to be compatible with the composed lexemes to the extent that is distributionally compatible with the lexeme that it is least compatible with. This corresponds to the multiset version of intersection. In particular, for all $\tau\in\xoverline{R}^*R^*$ and $w'\in V$: 
\begin{equation}
	\merge_{\MIN}{\set{\aptvar_1,\ldots,\aptvar_n}}(\tau,w') = \min_{1\le i\le n}{\aptvar_i(\tau,w')}
\label{eq:min}
\end{equation}
It is clear that the effectiveness of $\merge_{\MIN}$  increases as the size of $C$ grows, and that it would particularly benefit from distributional smoothing~\cite{Dagan_1994} which can be used to improve plausible co-occurrence coverage by inferring co-occurrences in the {\APT} for a lexeme $w$ based on the co-occurrences in the {\APT}s of distributionally similar lexemes.

An alternative to $\merge_{\MIN}$ is $\merge_{\SUMMING}$ where we determine distributional compatibility of a co-occurrence by aggregating across the distributional compatibility of the co-occurrence for each of the lexemes being composed. In particular, for all $\tau\in(\xoverline{R}\cup R)^*$ and $w'\in V$: 
\begin{equation}
	\merge_{\SUMMING}{\set{\aptvar_1,\ldots,\aptvar_n}}(\tau,w') = \sum_{1\le i\le n}{\aptvar_i(\tau,w')}
\label{eq:sum}
\end{equation}
While this clearly achieves co-occurrence embellishment, whether  co-occurrence filtering is achieved depends on the weighting scheme being used. For example, if negative weights are allowed, then co-occurrence filtering can be achieved. 

There is one very important feature of {\APT} composition  that is a distinctive aspect of our proposal, and therefore worth dwelling on. In Section~\ref{sec:approach}, when discussing Figure~\ref{fig:treecontext}, we made reference to the notions of internal and external context. The internal context of a composed {\APT} is that part of the {\APT} that corresponds to the nodes in the dependency tree that generated the composed {\APT}. One might have expected that the only lexeme appearing at an internal node is the lexeme that appears at the corresponding node in the dependency tree. However, this is absolutely not the objective: at each node in the internal context, we expect to find a set of alternative lexemes that are, to varying degrees, distributionally compatible with that position in the {\APT}. We expect that a lexeme that is distributionally compatible with a substantial number of the lexemes being composed will result in a  distributional feature with non-zero weight in the vectorized {\APT}. There is, therefore, no distinction being made between internal and external nodes. This enriches the distributional representation of the contextualized lexemes, and overcomes the potential problem arising from the fact that as larger and larger units are composed, there is less and less external context around to characterize distributional meaning.

\section{Experiments}
\label{sec:empirical}

In this section we consider some empirical evidence in support of {\APT}s.    First, we consider some of the different ways in which {\APT}s can be instantiated.   Second, we present a number of case studies showing the disambiguating effect of {\APT} composition in adjective-noun composition.  Finally, we evaluate the model using the phrase-based compositionality benchmarks of \namecite{Mitchell_2008} and \namecite{Mitchell_2010}.

\subsection{Instantiating {\APT}s}
\label{sec:instantiating}

We have constructed {\APT} lexicons from three different corpora. 
\begin{itemize}
	\item {\tt clean\_wiki} is a corpus used for the case studies in \ref{sec:emp-disambiguation}.  This corpus is a cleaned 2013 Wikipedia dump \cite{Wilson_2015} which we have tokenised, part-of-speech-tagged, lemmatised and dependency-parsed using the Malt Parser \cite{Nivre_2004}.  This corpus contains approximately 0.6 billion tokens.
	\item {\tt BNC} is the British National Corpus.  It has been tokenised, POS-tagged, lemmatised and dependency-parsed as described in \namecite{Grefenstette_2013} and contains approximately 0.1 billion tokens.
	\item {\tt concat} is a concatenation of the ukWaC corpus \cite{Ferraresi_2008}, a mid-2009 dump of the English Wikipedia and the British National Corpus.  This corpus has been tokenised, POS-tagged, lemmatised and dependency-parsed as described in \namecite{Grefenstette_2013} and contains about 2.8 billion tokens. 
\end{itemize}   

Having constructed lexicons, there are a number of hyperparameters to be explored during composition.  First there is the composition operation itself.  We have explored variants which take a union of the features such as {\tt add} and {\tt max} and variants which take an intersection of the features such as {\tt mult}, {\tt min} and {\tt intersective\_add}, where {\tt intersective\_add}$(a,b) = a+b$ iff $a > 0$ and $b > 0$; $0$ otherwise.
 
Second, the {\APT} theory is agnostic to the type or derivation of the weights which are being composed.  The weights in the elementary {\APT}s can be counts, probabilities, or some variant of PPMI or other association function.  Whilst it is generally accepted that the use of some association function such as PPMI is normally beneficial in the determination of lexical similarity, there is a choice over whether these weights should be seen as part of the representation of the lexeme, or as part of the similarity calculation.   In the instantiation which we refer to as as {\tt compose\_first}, {\APT} weights are probabilities.  These are composed and transformed to PPMI scores before computing cosine similarities.  In the instantiation which we refer to as {\tt compose\_second}, {\APT} weights are PPMI scores.

There are  a number of modifications that can be made to the standard PPMI calculation.  First, it is common \cite{Levy_2015} to delete rare words when building co-occurrence vectors.  Low frequency features contribute little to similarity calculations because they co-occur with very few of the targets.  Their inclusion will tend to reduce similarity scores across the board, but have little effect on ranking.  Filtering, on the other hand,   improves efficiency.  In other experiments, we have found that a feature frequency threshold of $1000$ works well.  On a corpus the size of Wikipedia ($~1.5$ billion tokens), this leads to a feature space for nouns of approximately $80,000$ dimensions (when including only first-order paths) and approximately $230,000$ dimensions (when including paths up to order $2$).

\namecite{Levy_2015} also showed that the use of context distribution smoothing ({\tt cds}), $\alpha = 0.75$, can lead to performance comparable with state-of-the-art word embeddings on word similarity tasks.  

\[\pmia{w}{\tau}{w'}{\alpha} = \log \frac{\cooccurs{\tco{w}{w'}{\tau}}\,\cooccurs{\tco{*}{*}{\tau}}^{\alpha}}{\cooccurs{\tco{w}{*}{\tau}}\,\cooccurs{\tco{*}{w'}{\tau}}^{\alpha}}\]

\namecite{Levy_2015} further showed that using shifted PMI, which is analogous to the use of negative sampling in word embeddings, can be advantageous.  When shifting PMI, all values are shifted down by $\log k$ before the threshold is applied.
   
 \[\sppmi{w}{\tau}{w'}= \max{ (\pmi{w}{\tau}{w'}- \log{k}, 0)}\]

Finally, there are many possible options for the path weighting function $\phi(\tau,w)$.  These include the path probability $\pcond{\tau}{w}$ as discussed in Section \ref{sec:similarity}, constant path weighting, and inverse path length or harmonic function (which is equivalent to the dynamic context window used in many neural implementations such as GloVe~\cite{Pennington_2014}).  

\subsection{Disambiguation}
\label{sec:emp-disambiguation}

Here we consider the differences between using aligned and unaligned {\APT} representations as well as the differences between using $\merge_{\SUMMING}$ and {$\merge_{\MIN}$} when carrying out adjective-noun (AN) composition.    From the {\tt clean\_wiki} corpus described in Section \ref{sec:instantiating}, a small number of high frequency nouns were chosen which are ambiguous or broad in meaning together with potentially disambiguating adjectives.   We use the {\tt compose\_first} option described above where composition is carried out on {\APT}s containing probabilities.

\[\weight{w}{\tau}{w'} = \frac{\cooccurs{\tco{w}{w'}{\tau}}}{\cooccurs{\tco{w}{*}{*}}}\]

The closest distributional neighbours of the individual lexemes before and after composition with the disambiguating adjective are then examined.  In order to calculate similarities, contexts are weighted using the variant of PPMI advocated by \namecite{Levy_2015} where{\tt cds} is applied with $\alpha=0.75$.  However, no shift is applied to the PMI values since we have found shifting to have little or negative effect when working with relatively small corpora.   Similarity is then computed using the standard cosine measure.   For illustrative purposes the top ten neighbours of each word or phrase are shown, concentrating on ranks rather than absolute similarity scores.  

\begin{table}[ht]
\centering
\begin{tabular}{|p{1.9cm}||p{2.3cm}|p{2.3cm}||p{2.3cm}|p{2.3cm}|}
\hline
&\multicolumn{2}{|c||}{Aligned $\merge_{\SUMMING}$}&\multicolumn{2}{|c|}{Unaligned $\merge_{\SUMMING}$}\\
\hline
shoot&green shoot&six-week shoot&green shoot&six-week shoot\\
\hline
shot \qquad\qquad leaf \qquad shooting \qquad fight\qquad \qquad scene \qquad video\qquad\qquad tour\qquad footage\qquad  interview\qquad  flower
&shoot \qquad\qquad {\bf leaf} \qquad  {\bf flower} \qquad {\bf fruit}\qquad\qquad  orange  \qquad{\bf tree}\qquad\qquad  color \qquad\qquad shot \qquad colour \qquad cover
&shoot \qquad\qquad {\bf tour} \qquad\qquad shot \qquad\qquad {\bf break}  \qquad{\bf session} \qquad {\bf show} \qquad shooting  {\bf concert}  {\bf interview} \qquad leaf
&shoot \qquad\qquad shot \qquad\qquad {leaf} \qquad shooting \qquad fight\qquad\qquad  scene\qquad  video\qquad\qquad  tour \qquad\qquad {flower} \qquad footage
&shoot \qquad\qquad shot\qquad\qquad  shooting \qquad leaf \qquad\qquad\qquad scene  \qquad{\bf video} \qquad fight \qquad {footage} \qquad {\bf photo}  {interview}\\
\hline
\end{tabular}
\vspace*{10pt}

\caption{Neighbours of uncontextualised shoot/N compared to shoot/N in the contexts of green/J and six-week/J, using $\merge_{\SUMMING}$ with aligned and unaligned representations}
\label{tab:shoot-sum}
\end{table} 

\begin{table}[ht]
\centering
\begin{tabular}{|p{1.7cm}||p{2.2cm}|p{2.6cm}||p{2.3cm}|p{2.3cm}|}
\hline
&\multicolumn{2}{|c||}{Aligned {$\merge_{\MIN}$}}&\multicolumn{2}{|c|}{Unaligned {$\merge_{\MIN}$}}\\
\hline
shoot&green shoot&six-week shoot&green shoot&six-week shoot\\
\hline
shot \qquad\qquad leaf \qquad shooting \qquad fight \qquad scene\qquad  video \qquad tour \qquad flower  \qquad footage \qquad interview
&shoot \qquad\qquad\qquad {\bf leaf} \qquad\qquad\qquad {\bf fruit}\qquad\qquad  {\bf stalk} \qquad {\bf flower}\qquad  {\bf twig} \qquad {\bf sprout} \qquad\qquad {\bf bud} \qquad\qquad {\bf shrub} \qquad\qquad {\bf inflorescence}
&shoot \qquad\qquad {\bf photoshoot}  {\bf taping} \qquad\qquad {\bf tour} \qquad\qquad {\bf airing} \qquad {\bf rehearsal}  {\bf broadcast} \qquad {\bf session}\qquad\qquad  q\&a \qquad\qquad\qquad {\bf post-production}
&shoot \qquad\qquad pyrite \qquad plosive \qquad handlebars \qquad annual \qquad roundel \qquad affricate \qquad phosphor  \qquad connections  reduplication
&e/f \qquad\qquad uemtsu \qquad\qquad confederations  shortlist \qquad all-ireland \qquad dern \qquad\qquad gerwen \qquad tactics  backstroke \qquad\qquad gabler\\
\hline
\end{tabular}
\vspace*{10pt}

\caption{Neighbours of uncontextualised shoot/N compared to shoot/N in the contexts of green/J and six-week/J, using {$\merge_{\MIN}$} with aligned and unaligned representations}
\label{tab:shoot-min}
\end{table}

Table \ref{tab:shoot-sum} illustrates what happens when {$\merge_{\SUMMING}$} is used to merge aligned and unaligned {\APT} representations when the noun \emph{shoot} is placed in the contexts of \emph{green} and {\emph{six-week}. Boldface is used in the entries of compounds where a neighbour appears to be highly suggestive of the intended sense and where it has a rank higher or equal to its rank in the entry for the uncontextualised noun.  In this example, it is clear that merging the unaligned {\APT} representations provides very little disambiguation of the target noun.  This is because typed co-occurrences for an adjective mostly belong in a different space to typed co-occurrences for a noun.  Addition of these spaces leads to significantly lower absolute similarity scores, but little change in the ranking of neighbours.  Whilst we only show one example here, this observation appears to hold true whenever words with different part of speech tags are composed.    Intersection of these spaces via {$\merge_{\MIN}$} generally leads to substantially degraded neighbours, often little better than random, as illustrated by Table~\ref{tab:shoot-min}.

On the other hand when {\APT}s are correctly aligned and merged using {$\merge_{\SUMMING}$}, we see the disambiguating effect of the adjective.  A \emph {green shoot} is more similar to \emph{leaf}, \emph{flower}, \emph{fruit} and \emph{tree}. 
A \emph{six-week shoot} is more similar to \emph{tour}, \emph{session}, \emph{show} and \emph{concert}.  This disambiguating effect is even more apparent when {$\merge_{\MIN}$} is used to merge the {\APT} representations (see Table~\ref{tab:shoot-min}).

{\small
\begin{table}
\centering
\begin{tabular}{|p{2.0cm}||p{2.3cm}|p{2.5cm}||p{2.3cm}|p{2.5cm}|}
\hline
&\multicolumn{2}{|c||}{Aligned {$\merge_{\SUMMING}$}}&\multicolumn{2}{|c|}{Aligned {$\merge_{\MIN}$}}\\
\hline
group&musical group&ethnic group&musical group&ethnic group\\
\hline
group organization organisation company community corporation unit movement association society
&group \qquad company \qquad {\bf band} \qquad\qquad{\bf music} movement community society corporation category  association
&group  organization  organisation  {\bf community}  company  movement  society \qquad\qquad {\bf minority}  \qquad\qquad unit \qquad  \qquad entity
&group  \qquad {\bf band} \qquad {\bf troupe}  {\bf ensemble}  {\bf artist} \qquad\qquad {\bf trio} \qquad\qquad  genre \qquad {\bf music}\qquad\qquad  {\bf duo} \qquad\qquad {\bf supergroup}
&group  {\bf community}  organization  grouping \qquad sub-group  {\bf faction}  {\bf ethnicity}  {\bf minority}  organisation  {\bf tribe}\\
\hline
body&human body&legislative body&human body&legislative body\\
\hline
body \qquad board \qquad organization  entity \qquad skin \qquad\qquad head  organisation \qquad  structure  council \qquad eye
&body  organization  structure \qquad entity\qquad \mbox{organisation} skin \qquad\qquad {\bf brain}  \qquad\qquad	{\bf eye} \qquad\qquad object \qquad {\bf organ}
&body \qquad\qquad {\bf council} \qquad {\bf committee}  board \qquad {\bf authority} {\bf  assembly}  organisation  {\bf agency}  {\bf commission}  entity
&body \qquad {\bf organism}  organization  entity \qquad {\bf embryo} \qquad {\bf brain}  community  {\bf organelle}  institution  {\bf cranium}
&body \qquad\qquad {\bf  council} \qquad {\bf committee}  board \qquad {\bf legislature}  {\bf secretariat}  {\bf authority}  {\bf assembly}  {\bf power} \qquad  {\bf office} \\
\hline
work&social work&literary work&social work&literary work\\
\hline
study  \qquad project \qquad book  \qquad activity \qquad effort  \qquad publication  job \qquad \qquad program  writing \qquad piece
&work \qquad activity \qquad study \qquad project  program  practice  {\bf development}  aspect \qquad book \qquad effort
&work \qquad\qquad {\bf book} \qquad\qquad study \qquad {\bf novel} \qquad project  \qquad \mbox{{\bf publication}}  {\bf text} \qquad\qquad {\bf literature} \qquad {\bf story} \qquad\qquad {\bf writing}
&work  \qquad\qquad research \qquad study \qquad\qquad writings  endeavour  project  discourse  topic \qquad {\bf development}  {\bf teaching}
&work \qquad\qquad  {\bf writings}  {\bf treatise} \qquad\qquad {\bf essay} \qquad {\bf poem} \qquad\qquad  book \qquad\qquad {\bf novel} \qquad {\bf monograph}  {\bf poetry} \qquad  {\bf writing}\\
\hline
field&athletic field&magnetic field&athletic field&magnetic field\\
\hline
facility \qquad stadium  \qquad area \qquad complex \qquad ground \qquad pool \qquad base \qquad space  \qquad centre \qquad park
&field \qquad\qquad facility  {\bf stadium}  {\bf gymnasium}  {\bf basketball}  {\bf sport} \qquad {\bf center}\qquad  {\bf softball} \qquad {\bf gym} \qquad {\bf arena} 
&field \qquad\qquad {\bf component}  stadium \qquad\qquad facility \qquad track \qquad ground \qquad system \qquad complex \qquad {\bf parameter}  pool 
&field  \qquad\qquad {\bf gymnasium}  {\bf fieldhouse}  {\bf stadium}  \qquad{\bf gym} \qquad  {\bf arena} \qquad {\bf rink} \qquad {\bf softball}  cafeteria  {\bf ballpark}
&field \qquad\qquad {\bf wavefunction}  \mbox{{\bf spacetime}}  \qquad {\bf flux} \qquad\qquad subfield \qquad\qquad \mbox{\bf perturbation}  {\bf vector}\qquad\qquad  {\bf e-magnetism}  formula\_8 \qquad {\bf scalar}\\\hline
\end{tabular}
\vspace*{10pt}

\caption{Distributional Neighbors using {$\merge_{\SUMMING}$} vs {$\merge_{\MIN}$} (e-magnetism = electro-magnetism)}
\label{tab:summin}
\end{table}
}

Table \ref{tab:summin} further illustrates the difference between using {$\merge_{\SUMMING}$} and {$\merge_{\MIN}$} when composing aligned {\APT} representations.  Again, boldface is used in the entries of compounds where a neighbour appears to be highly suggestive of the intended sense and where it has a rank higher or equal to its rank in the entry for the uncontextualised noun.  In these examples, we can see that both {$\merge_{\SUMMING}$} and $\merge_{\MIN}$ appear to be effective in carrying out some disambiguation.  Looking at the example of \emph{musical group}, both {$\merge_{\SUMMING}$} and $\merge_{\MIN}$ increase the relative similarity of \emph{band} and {\emph{music} to \emph{group} when it is contextualised by \emph{musical}.  However, $\merge_{\MIN}$ also leads to a number of other words being selected as neighbours which are closely related to the musical sense of group e.g. \emph{troupe}, \emph{ensemble} and \emph{trio}.  This is not the case when {$\merge_{\SUMMING}$} is used --- the other neighbours still appear related to the general meaning of \emph{group}.  This trend is also seen in some of the other examples such as \emph{ethnic group}, \emph{human body} and \emph{magnetic field}.  Further, even when {$\merge_{\SUMMING}$} leads to the successful selection of a large number of sense specific neighbours, e.g. see \emph{literary work}, the neighbours selected appear to be higher frequency, more general words than when $\merge_{\MIN}$ is used.  

The reason for this is likely to be the effect that each of these composition operations has on the number of non-zero dimensions in the composed representations.  Ignoring the relatively small effect the feature association function may have on this, it is obvious that {$\merge_{\SUMMING}$} should increase the number of non-zero dimensions whereas $\merge_{\MIN}$ should decrease the number of non-zero dimensions.  In general, the number of non-zero dimensions is highly correlated with frequency, which makes composed representations based on {$\merge_{\SUMMING}$} behave like high frequency words and composed representations based on $\merge_{\MIN}$ behave like low frequency words.  Further, when using similarity measures based on PPMI, as demonstrated by \namecite{Weeds_2003b}, it is not unusual to find that the neighbours of high frequency entities (with a large number of non-zero dimensions) are other high frequency entities (also with a large number of non-zero dimensions).   Nor is it unusual to find that the neighbours of low frequency entities (with a small number of non-zero dimensions) are other low frequency entities (with a small number of non-zero dimensions).   \namecite{Weeds_2004} showed that frequency is also a surprisingly good indicator of the generality of the word.   Hence {$\merge_{\SUMMING}$} leads to more general neighbours and $\merge_{\MIN}$ leads to more specific neighbours.  

Finally,  note that whilst $\merge_{\MIN}$ has produced high quality neighbours in these examples where only two words are composed, using $\merge_{\MIN}$ in the context of the composition of an entire sentence would tend to lead to very sparse representations.  The majority of the internal nodes of the {\APT} composed using an intersective operation such as $\merge_{\MIN}$ must necessarily only include the lexemes actually used in the sentence.  $\merge_{\SUMMING}$ on the other hand will have added to these internal representations, suggesting similar words which might have been used in those contexts and giving rise to a rich representation which might be used to calculate sentence similarity.   Further, the use of PPMI, or some other similar form of feature weighting and selection, will mean that those internal (and external) contexts which are not supported by a majority of the lexemes in the sentence will tend to be considered insignificant and therefore will be ignored in similarity calculations.   By using shifted PPMI, it should be possible to further reduce the number of non-zero dimensions in a representation constructed using $\merge_{\SUMMING}$ which should also allow us to control the specificity/generality of the neighbours observed.

\subsection{Phrase-based Composition Tasks}

Here we look at the performance of one instantiation of the {\APT} framework on two benchmark tasks for phrase-based composition.  

\subsubsection{Experiment 1: the M\&L2010 dataset}
\label{sect:ml2010}

The first experiment uses the M\&L2010 dataset, introduced by \namecite{Mitchell_2010}, which contains human similarity judgements for adjective-noun (AN), noun-noun (NN) and verb-object (VO) combinations on a seven-point rating scale.  It contains 108 combinations in each category such as $\langle\mbox{\em social activity},\mbox{\em economic condition}\rangle$, $\langle\mbox{\em tv set},\mbox{\em bedroom window}\rangle$ and $\langle\mbox{\em fight war},\mbox{\em win battle}\rangle$.   This dataset has been used in a number of evaluations of compositional methods including \namecite{Mitchell_2010}, \namecite{Blacoe_2012}, \namecite{Turney_2012}, \namecite{Hermann_2013} and \namecite{Kiela_2014b}.  For example, \namecite{Blacoe_2012} show that multiplication in a simple distributional space (referred to here as an untyped VSM) outperforms the distributional memory (DM) method of \namecite{Baroni_2010} and the neural language model (NLM) method of \namecite{Collobert_2008}.  

Whilst often not explicit, the experimental procedure in most of this work would appear to be the calculation of Spearman's rank correlation coefficient $\rho$ between model scores and individual, non-aggregated, human ratings.  For example, if there are 108 phrase pairs being judged by 6 humans, this would lead to a dataset containing 648 data points.  The procedure is discussed at length in \namecite{Turney_2012}, who argues that this method tends to underestimate model performance.  Accordingly, Turney explicitly uses a different procedure where a separate Spearman's $\rho$ is calculated between the model scores and the scores of each participant.   These  coefficients are then averaged to give the performance indicator for each model.    Here, we report results using the original {\tt M\&L} method, see Table \ref{tab:ml2010ml}.  We found that using the Turney method scores were typically higher by $0.01$ to $0.04$. If model scores are evaluated against aggregated human scores, then the values of Spearman's $\rho$ tends to be still higher, typically $0.1$ to $0.12$ higher than the values reported here.

\begin{table}[ht]
\begin{tabular}{|l|r|r|r|r|}\hline
&\bf AN&\bf NN&\bf VO&\bf Average\\
\hline\hline
$\merge_{\MIN}$, $k=1$&-0.09&0.43&0.35&0.23\\
$\merge_{\MIN}$, $k=10$&NaN&0.23&0.26&0.16\\[3pt]
$\merge_{\SUMMING}$, $k=1$&0.47&0.37&0.40&0.41\\
$\merge_{\SUMMING}$, $k=10$&0.45&0.42&\bf 0.42&0.43\\\hline
\hline
untyped VSM, {\tt multiply} &0.46&0.49&0.37&\bf 0.44\\
\cite{Mitchell_2010} &&&& \\[2pt]
untyped VSM, {\tt multiply}  &\bf 0.48&\bf 0.50&0.35&\bf 0.44\\
\cite{Blacoe_2012} &&&&\\[2pt]
distributional memory (DM), {\tt add}&0.37&0.30&0.29&0.32\\ 
\cite{Blacoe_2012}&&&&\\[2pt]
neural language model (NLM), {\tt add}&0.28&0.26&0.24&0.26\\
\cite{Blacoe_2012}&&&&\\\hline\hline
humans &0.52&0.49&0.55&0.52\\
\cite{Mitchell_2010}&&&&\\\hline
\end{tabular}
\vspace*{10pt}
\caption{Results on the M\&L2010 dataset using the {\tt M\&L} method of evaluation. Values shown are Spearman's $\rho$.}
\label{tab:ml2010ml}
\end{table}

For this experiment, we have constructed an order 2 {\APT} lexicon for the {\tt BNC} corpus.  This is the same corpus used by \namecite{Mitchell_2010} and for the best performing algorithms in \namecite{Blacoe_2012}.  We note that the larger {\tt concat} corpus was used by \namecite{Blacoe_2012} in the evaluation of the DM algorithm \cite{Baroni_2010b}}.  We use the {\tt compose\_second} option described above where the elementary {\APT} weights are PPMI.   With regard to the different parameter settings in the PPMI calculation \cite{Levy_2015}, we tuned on a number of popular word similarity tasks: MEN~\cite{Bruni_2014}; WordSim-353~\cite{Finkelstein_2001}}; and SimLex-999~\cite{Hill_2015}.  In these tuning experiments, we found that context distribution smoothing gave mixed results.  However, shifting PPMI ($k=10$) gave optimal results across all of the word similarity tasks.  Therefore we report results here for vanilla PPMI (shift $k=1$) and shifted PPMI (shift $k=10$).  For composition, we report results for both $\merge_{\SUMMING}$ and $\merge_{\MIN}$.   Results are shown in Table~\ref{tab:ml2010ml}.  

For this task and with this corpus $\merge_{\SUMMING}$ consistently outperforms $\merge_{\MIN}$.    Shifting PPMI by $\log 10$  consistently improves results for $\merge_{\SUMMING}$, but has a large negative effect on the results for $\merge_{\MIN}$.  We believe that this is due to the relatively small size of the corpus.  Shifting PPMI reduces the number of non-zero dimensions in each vector which increases the likelihood of a zero intersection.  In the case of AN composition, all of the intersections were zero for this setting, making it impossible to compute a correlation.

Comparing these results with the state-of-the-art, we can see that $\merge_{\SUMMING}$ clearly outperforms DM and NLM as tested by \namecite{Blacoe_2012}.  This method of composition is also achieving close to the best results in \namecite{Mitchell_2010} and \namecite{Blacoe_2012}.  It is interesting to note that our model does substantially better than the state-of-the-art on verb-object composition, but is considerably worse at noun-noun composition.  Exploring why this is so is a matter for further research.  We have undertaken experiments with a larger corpus and a larger range of hyper-parameter settings which indicate that the performance of the {\APT} models can be increased significantly. However, these results are not presented here, since an equatable comparison with existing models would require a similar exploration of the hyper-parameter space across all models being compared.

\subsubsection{Experiment 2: the M\&L2008 dataset}
\label{sect:ml2008}

The second experiment uses the M\&L2008 dataset, introduced by \namecite{Mitchell_2008}, which contains pairs of intransitive sensitives together with human judgments of similarity. The dataset contains 120 unique subject, verb, landmark triples with a varying number of human judgments per item.  On average each triple is rated by 30 participants.  The task is to rate the similarity of the verb and the landmark given the potentially disambiguating context of the subject.  For example, in the context of the subject \emph{fire} one might expect \emph{glowed} to be close to \emph{burned} but not close to \emph{beamed}.  Conversely, in the context of the subject \emph{face} one might expect \emph{glowed} to be close to \emph{beamed} and not close to \emph{burned}. 
   
This dataset was used in the evaluations carried out by \namecite{Grefenstette_2013} and \namecite{Dinu_2013}.  These evaluations clearly follow the experimental procedure of Mitchell and Lapata and do not evaluate against mean scores.  Instead, separate points are created for each human annotator, as discussed in Section \ref{sect:ml2010}.  

The multi-step regression algorithm of \namecite{Grefenstette_2013} achieved $\rho=0.23$ on this dataset.  In the evaluation of \namecite{Dinu_2013}, the lexical function algorithm, which learns a matrix representation for each functor and defines composition as matrix-vector multiplication,  was the best performing compositional algorithm at this task.  With optimal parameter settings, it achieved around $\rho=0.26$.  In this evaluation, the full additive model of \namecite{Guevara_2010}  achieved $\rho < 0.05$.  

In order to make our results directly comparable with these previous evaluations, we have used the same corpus to construct our {\APT} lexicons, namely the {\tt concat} corpus described in Section \ref{sec:instantiating}.  Otherwise, the {\APT} lexicon was constructed as described in Section \ref{sect:ml2010}.    As before note that $k=1$ in shifted PPMI is equivalent to not shifting PPMI.  Results are shown in Table~\ref{tab:ml2008ml}.   

\begin{table}[ht]
\begin{tabular}{|l|r|}
\hline
$\merge_{\MIN}$, $k=1$&0.23\\
$\merge_{\MIN}$, $k=10$&0.13\\[3pt]
$\merge_{\SUMMING}$, $k=1$&0.20\\
$\merge_{\SUMMING}$, $k=10$&\bf{0.26}\\
\hline\hline
multi-step regression&0.23\\
\namecite{Grefenstette_2013}&\\[2pt]
lexical function&0.23--{\bf 0.26}\\
\namecite{Dinu_2013}&\\[2pt]
untyped VSM, {\tt mult}&0.20-0.22\\
\namecite{Dinu_2013}&\\[2pt]
full additive&0--0.05\\
\namecite{Dinu_2013}&\\\hline\hline
humans&0.40\\
\namecite{Mitchell_2008}&\\\hline
\end{tabular}
\caption{Results on the M\&L2008 dataset. Values shown are Spearman's $\rho$.}
\label{tab:ml2008ml}
\end{table}

We see that $\merge_{\SUMMING}$ is highly competitive with the optimised lexical function model which was the best performing model in the evaluation of \namecite{Dinu_2013}.  In that evaluation, the lexical function model achieved between 0.23 and 0.26 depending on the parameters used in dimensionality reduction.   Using vanilla PPMI, without any context distribution smoothing or shifting, $\merge_{\SUMMING}$ achieves $\rho=0.20$, which is less than $\merge_{\MIN}$.  However, when using shifted PPMI as weights, the best result is $0.26$.  The shifting of PPMI means that contexts need to be more surprising in order to be considered as features.  This makes sense when using an additive model such as $\merge_{\SUMMING}$.

We also see that at this task and using this corpus $\merge_{\MIN}$ performs relatively well.  Using vanilla PPMI, without any context distribution smoothing or shifting, it achieves $\rho=0.23$ which equals the performance of the multi-step regression algorithm \namecite{Grefenstette_2013}.  Here, however, shifting PPMI has a negative impact on performance.  This is largely due to the intersective nature of the composition operation --- if shifting PPMI removes a feature from one of the unigram representations, it cannot be recovered during composition.  

\section{Related Work}
\label{sec:related}

Our work brings together two strands usually treated as separate though related problems: representing phrasal meaning by creating distributional representations through composition; and representing word meaning in context by modifying the distributional representation of a word.  In common with some other work on lexical distributional similarity, we use a typed co-occurrence space.  However, we propose the use of higher-order grammatical dependency relations to enable the representation of phrasal meaning and the representation of word meaning in context.  

\subsection{Representing Phrasal Meaning}

The problem of representing phrasal meaning has traditionally been tackled by taking vector representations for words \cite{Turney_2010} and combining them using some function to produce a data structure that represents the phrase or sentence. Mitchell and Lapata (2008, 2010)\nocite{Mitchell_2008,Mitchell_2010} found that simple additive and multiplicative functions applied to proximity-based vector representations were no less effective than more complex functions when performance was assessed against human similarity judgements of simple paired phrases. 

The word embeddings learnt by the continuous bag-of-words model (CBOW) and the continuous skip-gram model proposed by \nocite{Mikolov_2013,Mikolov_2013} Mikolov et al. (2013a, 2013b) are currently among the most popular forms of distributional word representations.  Whilst using a neural network architecture, the intuitions behind such distributed representations of words are the same as in traditional distributional representations.  As argued by \nocite{Pennington_2014} Pennington et al. (2014), both count-based and prediction-based models probe the underlying corpus co-occurrences statistics.  For example, the CBOW architecture predicts the current word based on context (which is viewed as a bag-of-words) and the skip-gram architecture predicts surrounding words given the current word.    \nocite{Mikolov_2013c} Mikolov et al. (2013c) showed that it is possible to use these models to efficiently learn low-dimensional representations for words which appear to capture both syntactic and semantic regularities.   \namecite{Mikolov_2013b} also demonstrated the possibility of composing skip-gram representations using addition.  For example, they found that adding the vectors for \emph{Russian} and \emph{river} results in a very similar vector to the result of adding the vectors for \emph{Volga} and \emph{river}.   This is similar to the multiplicative model of \namecite{Mitchell_2008} since the sum of two skip-gram word vectors is related to the product of two word context distributions.  

Whilst our model shares with these the use of vector addition as a composition operation, the underlying framework is very different.  Specifically, the actual vectors added depend not just on the form of the words but also their grammatical relationship within the phrase or sentence.  This means that the representation for, say, \emph{glass window} is not equal to the representation of \emph{window glass}.  The direction of the $\ctA{nn}$ relationship between the words leads to a different alignment of the {\APT}s and consequently a different representation for the phrases.

There are other approaches which incorporate theoretical ideas from formal semantics and machine learning, use syntactic information, and specialise the data structures to the task in hand. 
For adjective-noun phrase composition, \namecite{Baroni_2010} and \namecite{Guevara_2010} borrowed from formal semantics the notion that an adjective acts as a modifying function on the noun. They represented a noun as a vector, an adjective as a matrix, which could be induced from pairs of nouns and adjective noun phrases, and composed the two using matrix-by-vector multiplication to produce a vector for the noun phrase. Separately, \namecite{Coecke_2011} proposed a broader compositional framework that incorporated from formal semantics the notion of function application derived from syntactic structure \cite{Montague_1970,Lambek_1999}. These two approaches were subsequently combined and extended to incorporate simple transitive and intransitive sentences, with functions represented by tensors, and arguments represented by vectors \cite{Grefenstette_2013}.

The MV-RNN model of \namecite{Socher_2012} broadened the \namecite{Baroni_2010} approach; all words, regardless of part-of-speech, were modelled with both a vector and a matrix. This approach also shared features with \namecite{Coecke_2011} in using syntax to guide the order of phrasal composition. This model, however, was made much more flexible by requiring and using task-specific labelled training data to create task-specific distributional data structures, and by allowing non-linear relationships between component data structures and the composed result. The payoff for this increased flexibility has come with impressive performance in sentiment analysis \cite{Socher_2012,Socher_2013b}.

However, whilst these approaches all pay attention to syntax, they all require large amounts of training data.  For example, running regression models to accurately predict the matrix or tensor for each individual adjective or verb requires a large number of exemplar compositions containing that adjective or verb.  Socher's MV-RNN model further requires task-specific labelled training data.     Our approach, on the other hand,  is purely count-based and directly aggregates information about each word from the corpus. 

Other approaches have been proposed. \nocite{Clarke_2007,Clarke_2012} Clarke (2007, 2012) suggested a context-theoretic semantic framework, incorporating a generative model that assigned probabilities to arbitrary word sequences. This approach shared with \namecite{Coecke_2011} an ambition to provide a bridge between compositional distributional semantics and formal logic-based semantics. In a similar vein, \namecite{Garrette_2011} combined word-level distributional vector representations with logic-based representation using a probabilistic reasoning framework.  \namecite{Lewis_2013} also attempted to combine distributional and logical semantics by learning a lexicon for CCG (Combinatory Categorial Grammar \cite{Steedman_2000}) which first maps natural language to a deterministic logical form and then performs a distributional clustering over logical predicates based on arguments.  The CCG formalism was also used by \namecite{Hermann_2013} as a means for incorporating syntax-sensitivity into vector space representations of sentential semantics based on recursive auto-encoders \nocite{Socher_2011,Socher_2011b} (Socher et al. (2011a, 2011b)).   They achieved this by representing each combinatory step in a CCG parse tree with an auto-encoder function, where it is possible to parameterise both the weight matrix and bias on the combinatory rule and the CCG category.

 \namecite{Turney_2012} offered a model that incorporated assessments of word-level semantic relations in order to determine phrasal-level similarity.  This work uses two different word-level distributional representations to encapsulate two types of similarity, and captures instances where the components of a composed noun phrase bore similarity to another word through a mix of those similarity types.  Crucially, it views similarity of phrases as a function of the similarities of the components and does not attempt to derive modified vectors for phrases or words in context.  \namecite{Dinu_2012} also compared computing sentence similarity via additive compositional models with an alignment-based approach, where sentence similarity is a function of the similarities of component words, and simple word overlap.  Their results showed that a model based on a mixture of these approaches outperformed all of the individual approaches on a number of textual entailment datasets.

\subsection{Typed Co-occurrence Models}

In untyped co-occurrence models, such as those considered by Mitchell and Lapata (2008, 2010) \nocite{Mitchell_2008,Mitchell_2010}, co-occurrences are simple, untyped pairs of words which co-occur together (usually within some window of proximity but possibly within some grammatical relation).  The lack of typing makes it possible to compose vectors through addition and multiplication.  However, in the computation of lexical distributional similarity using grammatical dependency relations, it has been typical  \cite{Lin_1998b,Lee_1999,Weeds_2005}  to consider the \emph{type} of a co-occurrence (for example, does \emph{dog} occur with \emph{eat} as its direct object or its subject?) as part of the feature space.  The distinction between vector spaces based on untyped and typed co-occurrences was formalised by \namecite{Pado_2007} and \namecite{Baroni_2010b}.  In particular, \namecite{Baroni_2010b} showed that typed co-occurrences based on grammatical relations were better than untyped co-occurrences for distinguishing certain semantic relations.  However, as shown by \namecite{Weeds_2014}, it does not make sense to compose typed features based on first-order dependency relations through multiplication and addition, since the vector spaces for different parts of speech are largely non-overlapping.

\namecite{Pado_2007} constructed features using higher-order grammatical dependency relations.  They defined a path through a dependency tree in terms of the node words.  This allowed words which are only indirectly related within a sentence to be considered as co-occurring.  For example, in \emph{a lorry carries apples}, there is a path of length 2 between the nouns \emph{lorry} and \emph{apples} via the node \emph{carry}.  However, they also used a word-based basis mapping which essentially reduces all of the salient grammatical paths to untyped co-occurrences.  Given the paths $\langle\mbox{\em lorry}, \mbox{\em carry}\rangle$ and $\langle\mbox{\em lorry}, \mbox{\em carry}, \mbox{\em apples}\rangle$ for \emph{lorry}, these would be mapped to the basis elements \emph{carry} and \emph{apples} respectively.

\subsection{Representing Word Meaning in Context}

A long-standing topic in distributional semantics has been the modification of a canonical representation of a lexeme's meaning to reflect the context in which it is found.  Typically, a canonical vector for a lexeme is estimated from all corpus occurrences and the vector then modified to reflect the instance context 
\cite{Lund_1996,Erk_2008,Mitchell_2008,Thater_2009,Thater_2010,Thater_2011,VanDeCruys_2011,Erk_2012}. 

As described in Mitchell and Lapata (2008, 2010), lexeme vectors have typically been modified using simple additive and multiplicative compositional functions. Other approaches, however, share with our proposal the use of syntax to drive modification of the distributional representation \cite{Erk_2008,Thater_2009,Thater_2010,Thater_2011}.   

\namecite{Erk_2008} introduced a structured vector space model of word meaning that computes the meaning of a word in the context of another word via selectional preferences.  This approach was shown to work well at ranking paraphrases taken from the SemEval-2007 lexical substitution task \cite{McCarthy_2007}.     In the Erk \& Pad\'{o} approach, the meaning of \emph{ball} in the context of the phrase \emph{catch ball} is computed by combining the lexical vector for \emph{ball} with the object preference vector of \emph{catch} i.e. things which can be \emph{caught}.   Whilst this approach is based on very similar intuitions to ours, it is in fact quite different.  The lexical vector which is modified is not the co-occurrence vector, as in our model, but a vector of neighbours computed from co-occurrences.   For example, the \emph{lexical} vector for \emph{catch} in the Erk \& Pad\'{o} approach might contain \emph{throw}, \emph{catch} and \emph{organise}.   These neighbours of \emph{catch} are then combined with verbs which have been seen with \emph{ball} in the direct object relation using vector addition or component-wise multiplication.  Thus, it is possible to carry out this approach with reference only to observed first order grammatical dependency relationship.  In their experiments, they used the ``dependency-based'' vector space of \namecite{Pado_2007} where target and context words are linked by a valid dependency path (i.e. not necessarily a single first-order grammatical relation).  However, higher-order dependency paths were purely used to provide extra contexts for target words, than would be seen in a traditional first-order dependency model, during the computation of neighbour sets.  Further, the Erk \& Pad\'{o} approach does not construct a representation of the phrase since this model is focussed on lexical disambiguation rather than composition and it is not obvious how one would carry out further disambiguations within the context of a whole sentence. 
  
More recently, \namecite{Thater_2011} used a similar approach but considered a broader range of operations for combining two vectors where individual vector components are reweighted.  Specifically, they found that reweighting vector components based on the distributional similarity score between words defining vector components and the observed context words led to improved performance at ranking paraphrases.

\namecite{Thater_2010} noted that vectors of two syntactically related words typically have different syntactic environments, making it difficult to combine information in the respective vectors.  They build on  \namecite{Thater_2009}, where the meaning of argument nouns was modelled in terms of the predicates they co-occur with (referred to as a \emph{first-order} vector) and the meaning of predicates in terms of \emph{second-order} co-occurrence frequencies with other predicates.  These predicate vectors can be obtained by adding argument vectors.  For example, the verb \emph{catch} will contain counts on the dimension for \emph{kick} introduced by the direct-object \emph{ball} and counts on the dimension for \emph{contract} introduced by the direct-object \emph{cold}.  In other words, like in the Erk \& Pad\'{o} approach, the vector for a verb can be seen as a vector of similar verbs, thus making this notion of \emph{second-order} dependency compatible with that used in work on word sense discrimination \cite{Schutze_1998} rather than referring to second-order (or higher order) grammatical dependencies as in this work.  Contextualisation can then be achieved by multiplication of a second-order predicate vector with a first-order argument vector since this selects the dimensions which are common to both.  \namecite{Thater_2010} presented a more general model where every word is modelled in terms of first-order and second-order co-occurrences and demonstrate high performance at ranking paraphrases.

\section{Directions for Future Work}
\label{sec:future}

\subsection{Representations}

There are a number of apparent limitations of our approach that are simply a reflection of our decision to adopt dependency-based syntactic analysis.

First, surface disparities in syntactic structure (e.g. active versus passive tense formations, compound sentence structures) will disrupt sentence-level comparisons using a simple {\APT} structure based on surface dependency relations, but this can be addressed, for example, by syntax-based pre-processing. The {\APT} approach is agnostic in this regard. 

Second, traditional dependency parsing does not distinguish between the order of modifiers.  Hence the phrases  \emph{happiest blonde person} and \emph{blonde happiest person} receive the same dependency representation and therefore also the same semantic representation.  However, we believe that our approach is flexible enough to be able to accommodate a more sensitive grammar formalism which does allow for distinctions in modifier scope to be made if an application demands it.  In future work we intend to look at other grammar formalisms including CCG~\cite{Steedman_2000}.

By proposing a count-based method for composition we are bucking the growing trend of working with prediction-based word embeddings.  Whilst there has been initial evidence~\cite{Baroni_2014} that prediction-based methods are superior to count-based methods at the lexeme level e.g. for synonym detection and concept categorisation, it has also been shown~\cite{Levy_2014c} that the skip-gram model with negative sampling as introduced in \namecite{Mikolov_2013} is equivalent to implicit factorisation of the PPMI matrix.  \namecite{Levy_2015} also demonstrated how traditional count-based methods could be improved by transferring hyperparameters used by the prediction-based methods (such as context distribution smoothing and negative sampling).  This led to the count-based methods outperforming the prediction-based methods on a number of word similarity tasks.  A next step for us is to take the lessons learnt from work on word embeddings and find a way to produce lower dimensionality {\APT} representations without destroying the necessary structure which drives composition.  The advantages of this from a computational point of view are obvious.  It remains to be seen what effect the improved generalization also promised by dimensionality reduction will have on composition via {\APT}s.

By considering examples, we have seen that composition of {\APT}s using both union and intersection can lead to nearest neighbours which are clearly disambiguating.  On benchmark phrase-based composition tasks, the performance of union in  {\APT} composition is close to or equalling the state-of-the-art on those tasks.  However, we believe that the performance of intersection in {\APT} composition is currently limited by the impoverished nature of word representations based directly on corpus statistics.   Even given a very large corpus, there are always many plausible co-occurrences which have not been observed.  One possible solution, which we explore elsewhere, is to smooth the word representations using their distributional neighbours before applying an intersective composition operation. 

\subsection{Applications}

In Section \ref{sec:emp-disambiguation}, we demonstrated the potential for using {\APT}s to carry out word sense disambiguation / induction. Uncontextualised, elementary  {\APT}s typically contain a corpus-determined mixture of co-occurrences referencing different usages. The {\APT} generated by a dependency tree, however, provides contextualised lexeme representations where the weights have been adjusted by the influence of the contextual lexemes so that the co-occurrences relating to the \emph{correct} usage have been appropriately up-weighted, and the co-occurrences found in other circumstances down-weighted. In other words, {\APT} structures automatically perform word sense induction on lexeme-level representations which is demonstrable through the lexeme similarity measure. For example, we observed that the contextualised lexeme representation of \emph{body} in the {\APT} constructed by embedding it in the phrase \emph{human body} had a relatively high similarity to the uncontextualised representation of \emph{brain} and a relatively low similarity to \emph{council}, while the equivalent lexeme representation for \emph{body} embedded in the {\APT} constructed for the phrase \emph{legislative body} showed the reverse pattern.

One common criticism of distributional thesauruses is that they conflate different semantic relations into a single notion of similarity.  For example, when comparing representations based on grammatical dependency relations, the most similar word to an adjective such as \emph{hot} will usually be found to be its antonym \emph{cold}.  This is because \emph{hot} and \emph{cold} are both used to modify many of the same nouns.  However, if as in the {\APT} framework, the representation of \emph{cold} includes not only the direct dependents of \emph{cold}, but also the indirect dependents, e.g. verbs which co-occur with \emph{cold things}, it is possible that more differences between its representation and that of \emph{hot} might be found.  One would imagine that the things which are done to \emph{hot things} are more different to the things which are done to \emph{cold things} than they are to the things which are done to \emph{very warm things}.  Further, the examples in Section \ref{sec:emp-disambiguation} raises the possibility that different composition operations might be used to distinguish different semantic relations including hypernyms, hyponyms and co-hyponyms.  For example, $\merge_{\SUMMING}$ tends to lead to more general neighbours (e.g. hypernyms) and $\merge_{\MIN}$ tends to lead to more specific neighbours (e.g. hyponyms).  

Phrase-level or sentence-level plausibility measures offer the prospect of a continuous measure of the \emph{appropriateness} / \emph{plausibility} of a complete phrase or sentence, based on a combination of semantic and syntactic dependency relations.  {\APT}s offer a way to measure the plausibility of a lexeme when embedded in a dependency tree, suggesting that {\APT}s may be successfully employed in tackling sentence completion tasks, such as the Microsoft Research Sentence Completion Challenge~\cite{Zweig_2012}.  Here the objective is to identify the word that will fill out a partially completed sentence in the best possible way. For example, is \emph{flurried} or \emph{profitable} the best completion of the sentence below.
\begin{quote}
"Presently he emerged looking even more \emph{[flurried / profitable]} than before."
\end{quote}
We can compose the {\APT}s for the partially completed sentence.  Comparing the result with the elementary {\APT}s for each of the candidates should provide a good, direct measurement of which candidate is more plausible. An improved language model has implications for parsing, speech recognition and machine translation.

A central goal of compositional distributional semantics is to create a data structure that represents an entire phrase or sentence. The composed {\APT} for a dependency tree provides such a structure, but leaves open the question as to how this structure might be exploited for phrase-level or sentence-level semantic comparison.

\vspace*{3pt}

The first point to be made is that, unusually, we have available not only a representation of the whole dependency tree but also contextualised (vector) representations for the lexemes in the dependency tree. This makes available to us any analytical technique which requires separate analysis of lexical components of the phrase or sentence.  However, this leads to the problem of how to \emph{read} the structure at the global phrase/sentence-level. 

For similarity measures, one straightforward option would be to create a vector from the {\APT} anchored at the head of the phrase or sentence being considered. Thus the phrasal vector for \emph{a red rose} would be created taking the node containing \emph{rose} as the anchor. In other words, the vector representation of the phrase \emph{a red rose} will be the same as the contextualised representation of \emph{rose}. Similarly, the vector representation for the sentence \emph{he took the dog for a walk} will be the same as the contextualised representation of the verb \emph{took}.

Such a representation provides a continuous model of similarity (and meaning) at the phrasal-level and/or sentence-level. We  anticipate that vector comparisons of phrase or sentence-level vectors produced in this manner will provide some coherent numerical measure of distributional similarity. This approach should be useful for paraphrase recognition tasks.  For example, in order to identify good candidate paraphrases for questions in a question-answering task, \namecite{Berant_2014} employ a paraphrase model based on adding word embeddings constructed using the CBOW model of \nocite{Mikolov_2013}Mikolov et al. (2013).   Whilst the authors achieve state-of-the-art using a mixture of methods, a paraphrase model based on the addition of vectors of untyped co-occurrences alone cannot distinguish meanings where syntax is important.  For example, the sentences \emph{Oswald shot Kennedy} and \emph{Kennedy shot Oswald} would have the same representations.  On the other hand, {\APT} composition is syntax-driven and will provide a representation of each sentence which is sensitive to lexical meaning and syntax.  

Another advantage of using {\APT} composition in paraphrase recognition, over some other syntax-driven proposals, is that the same structure is used to represent words, phrases and sentences.  Provided the head node is of the same type of speech, words and phrases of different lengths can easily be compared within our model.  An adjective-noun compound such as \emph{male sibling} is directly comparable with the single noun \emph{brother}.  Further,  there is no need for there to be high similarity between aligned components of phrases or sentences.  For example, the phrase \emph{female scholar} can be expected to have a high similarity with the phrase \emph{educated woman}, in terms at least of their external contexts. 

\section{Conclusions}
\label{sec:conclusions}

This paper presents a new theory of compositional distributional semantics.   It employs a single structure, the {\APT}, which can represent the distributional semantics of lexemes, phrases and even sentences. By retaining higher-order grammatical structure in the representations of lexemes, composition captures mutual disambiguation and mutual generalisation of constituents.   {\APT}s allow lexemes and phrases to be compared in isolation or in context.  Further, we have demonstrated how one instantiation of this theory can achieve results which are very competitive with state-of-the-art results on benchmark phrase-based composition tasks.

As we have discussed, {\APT}s have a wide range of potential applications including word sense induction, word sense disambiguation, parse reranking, dependency parsing and language modelling more generally, and also paraphrase recognition.    Further work is required to gain an understanding of which instantiations of the theory are suited to each of these applications.

\section{Acknowledgements}

This work was funded by UK EPSRC project EP/IO37458/1 ``\emph{A Unified Model of Compositional and Distributional Compositional Semantics: Theory and Applications}''. We would like to thank all members of the DISCO project team. Particular thanks to Miroslav Batchkarov, Stephen Clark, Daoud Clarke, Roland Davis,  Bill Keller, Tamara Polajnar, Laura Rimell, Mehrnoosh Sadrzadeh, David Sheldrick and Andreas Vlachos.  We would also like to thank the anonymous reviewers for their helpful comments.

\bibliographystyle{fullname}
\bibliography{common}
\end{document}